\documentclass{article}
\pdfoutput=1

\PassOptionsToPackage{numbers, compress}{natbib}

\usepackage[final]{neurips_2024}

\usepackage[utf8]{inputenc} 
\usepackage[T1]{fontenc}    
\usepackage{url}            
\usepackage{booktabs}       
\usepackage{amsfonts}       
\usepackage{nicefrac}       
\usepackage{microtype}      

\usepackage{enumitem}
\usepackage{ulem}
\usepackage[dvipsnames,table]{xcolor}
\usepackage{graphicx}
\usepackage{amsmath}
\usepackage{amssymb}
\usepackage{booktabs}
\usepackage{makecell}
\usepackage{color}
\usepackage{pifont}
\usepackage{float}
\usepackage{wrapfig}
\usepackage[caption=false]{subfig}
\usepackage{bm}
\usepackage{multirow} 
\usepackage{adjustbox}
\usepackage{bbding}
\makeatletter
  \newcommand\figcaption{\def\@captype{figure}\caption}
  \newcommand\tabcaption{\def\@captype{table}\caption}
\makeatother

\newcommand\blfootnote[1]{%
  \begingroup
  \renewcommand\thefootnote{}%
  \footnotetext{#1}%
  \endgroup
}

\definecolor{annotation}{RGB}{0, 153, 0}
\definecolor{key_words}{RGB}{236, 0, 141}
\definecolor{RowColor}{rgb}{0.95, 0.95, 1}
\definecolor{cgray}{RGB}{220,220,220}
\definecolor{lightblue}{RGB}{163,199,235}
\definecolor{darkblue}{RGB}{0,76,153}
\definecolor{citegrey}{HTML}{75878a}
\definecolor{updatagreen}{RGB}{80,100,40}
\definecolor{updatagrey}{HTML}{686461}
\definecolor{citecolor}{HTML}{2980b9}
\definecolor{linkcolor}{HTML}{c0392b}

\newcommand{\pub}[1]{{\color{citegrey}{\tiny{[{#1}]}}}}
\newcommand{\updata}[1]{\fontsize{7.5pt}{1em}\selectfont{\color{gray}{~\textsl{\textbf{#1}}}}}

\usepackage{algorithm}
\usepackage{algorithmic}
\usepackage{newfloat}
\usepackage{listings}
\DeclareCaptionStyle{ruled}{labelfont=normalfont,labelsep=colon,strut=off} 
\floatstyle{ruled}
\newfloat{listing}{tb}{lst}{}
\floatname{listing}{Listing}

\newcommand{\cmark}{\textcolor{green}{\ding{51}}}%
\newcommand{\xmark}{\textcolor{red}{\ding{55}}}%

\usepackage[hidelinks,breaklinks=true,colorlinks,bookmarks=false,citecolor=citecolor,linkcolor=linkcolor]{hyperref}
\title{Target-Guided Adversarial Point Cloud Transformer Towards Recognition Against Real-world Corruptions}

\author{%
  \textbf{Jie Wang$^{1}$, Tingfa Xu$^{1, \dag}$, Lihe Ding$^{2}$,  Jianan Li$^{1,\dag}$} \\
  $^{1}$Beijing Institute of Technology \quad $^{2}$The Chinese University of Hong Kong\\
  \texttt{\url{https://github.com/Roywangj/APCT}}
}

\begin{document}

\maketitle
\blfootnote{$^{\dag}$Correspondence to: Jianan Li and Tingfa Xu.}

\begin{abstract}
Achieving robust 3D perception in the face of corrupted data presents an challenging hurdle within 3D vision research. Contemporary transformer-based point cloud recognition models, albeit advanced, tend to overfit to specific patterns, consequently undermining their robustness against corruption. In this work, we introduce the Target-Guided \textbf{A}dversarial \textbf{P}oint \textbf{C}loud \textbf{T}ransformer, termed \textbf{APCT}, a novel architecture designed to augment global structure capture through an adversarial feature erasing mechanism predicated on patterns discerned at each step during training. 
Specifically, APCT integrates an {Adversarial Significance Identifier} and a {Target-guided Promptor}. 
The {Adversarial Significance Identifier}, is tasked with discerning token significance by integrating global contextual analysis, utilizing a structural salience index algorithm alongside an auxiliary supervisory mechanism. 
The {Target-guided Promptor}, is responsible for accentuating the propensity for token discard within the self-attention mechanism, utilizing the value derived above, consequently directing the model attention towards alternative segments in subsequent stages.
By iteratively applying this strategy in multiple steps during training, the network progressively identifies and integrates an expanded array of object-associated patterns.
Extensive experiments demonstrate that our method achieves state-of-the-art results on multiple corruption benchmarks.
\end{abstract}

\vspace{-5mm}
\section{Introduction}
\vspace{-3mm}
3D point cloud recognition has garnered significant interest owing to its promising implications for robotics and autonomous driving. Prevailing techniques~\cite{qi2017pointnet, qi2017pointnet++, wang2021papooling, qianpointnext} have been predominantly designed and evaluated on clean data~\cite{modelnet40, chang2015shapenet}, overlooking the extensive corruptions present in real-world scenarios
arising from sensor inaccuracies and physical constraints and leading to suboptimal performance when models are exposed to such conditions. Therefore, enhancing the resilience of point cloud models against real-world corruption emerges as a paramount yet daunting task.

\begin{figure*}[t!]
  \centering
  \setlength{\abovecaptionskip}{1pt}  
  \setlength{\belowcaptionskip}{-5pt}
  \includegraphics[width=0.95\textwidth]{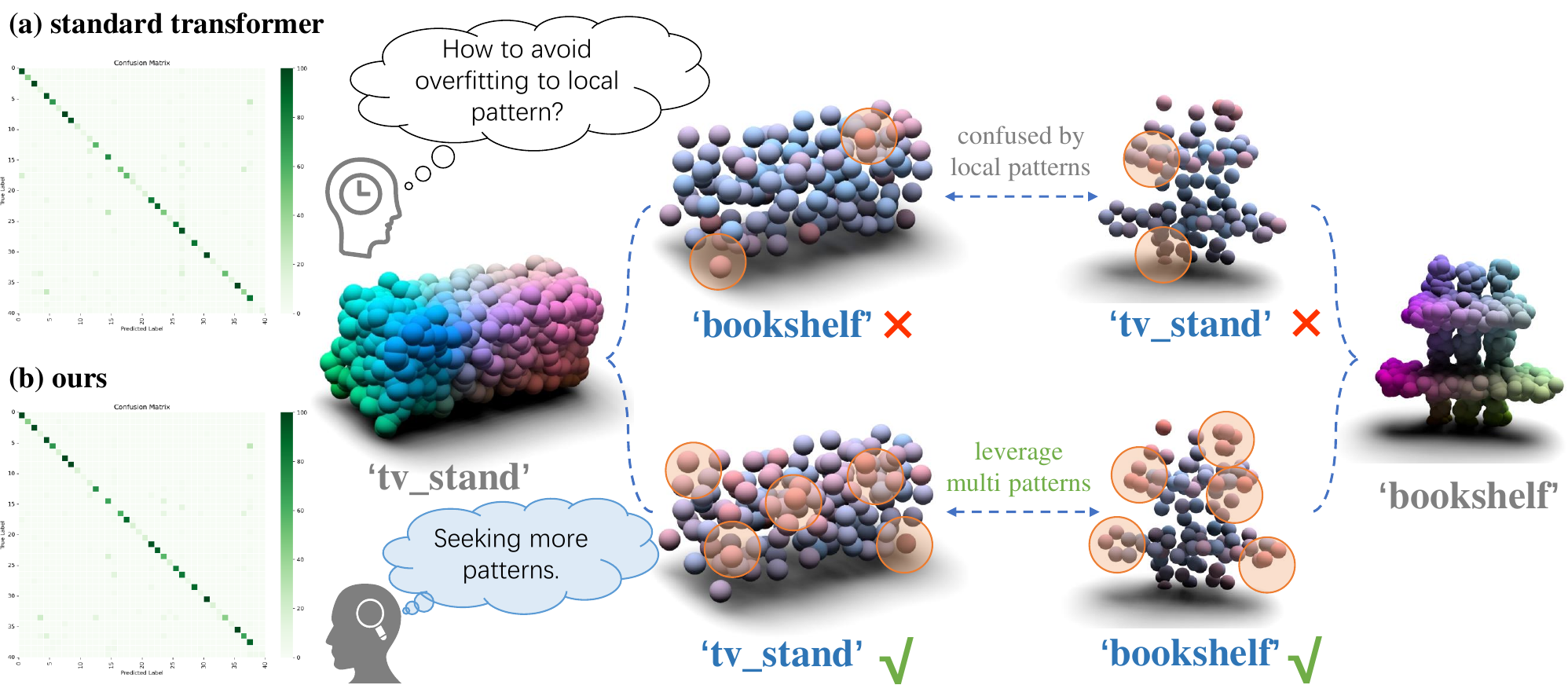}
  \caption{\textbf{Overall motivation}. We advocate for the model to broaden its attention to diverse patterns, mitigating the tendency to overfit to localized patterns. The left segment of the figure contrasts the confusion matrices of the standard transformer with our approach. The right portion showcases the performances of both the standard transformer and our methodology when confronted with objects exhibiting similar local patterns. Tokens with \textcolor{linkcolor}{high} / \textcolor{citecolor}{low} contributions to classification are in \textcolor{linkcolor}{red} / \textcolor{citecolor}{blue}, respectively. Standard transformer tends to overfit to localized patterns. 
  While our method, by modulating tokens with significant contributions, enables the model to garner features from a varied spectrum of target segments, thereby ensuring greater robustness.
  }
  \label{fig:motivation}
\vspace{-4mm}
\end{figure*}

Due to their elevated performance, existing transformer-based models~\cite{guo2021pct, ren2022benchmarking} have emerged as the mainstream choice. Despite their effectiveness on clean datasets, these methods generally falter when faced with corrupted data.
Upon closer examination, we discerned that prevailing models have a propensity to overfit to specific patterns~(Fig.~\ref{fig:motivation}\textcolor{linkcolor}{a}). Such patterns can degrade in the presence of real-world corruptions, undermining the model reliability. Relying solely on these localized patterns for predictions can be fragile, especially with corrupted data. Thus, we theorize that if the model is encouraged to extract features from a broader region of the object during training, gathering more diverse perception cues, it would be more resilient, as proved in Fig.~\ref{fig:motivation}\textcolor{linkcolor}{b}. This is because, even if some local patterns are compromised in corrupted data, the model could still source information from other intact areas to make accurate predictions.

To address the aforementioned limitation, we present a novel adversarial  representation learning method, named Target-Guided Adversarial Point Cloud Transformer(APCT), for robust 3D perception. This method  adversarially weakens the dominant patterns and gathers more sub-important perception cues during training on clean samples.
By progressive pattern excavation, the proposed approach  prompts the model to delve into objects and acquire a broader range of patterns, consequently elevating the robustness of the model. %

Specifically, we first partition and extract local geometries from the point cloud by a mini-PointNet~\cite{qi2017pointnet}, resulting in a subset of tokens that encode these features. The tokens are then fed into stacked transformer blocks with two core modules, namely the {Adversarial Significance Identifier} and the {Target-guided Promptor}. 
The former module utilizes a dominant feature index mechanism along with an auxiliary supervision, to discern significance of all tokens, thus integrating global contextual analysis.
Simultaneously, it signs a proportion of dominant tokens that are significant for  perception in the current phase.
Subsequently, the {Target-guided Promptor}, increases the dropping likelihood of token signed above during the self-attention process, thereby compelling the network to focus on less dominant tokens and to extract perceptive clues from alternative patterns. 
By iteratively engaging in such an adversarial process, the network gradually excavates and assimilates an extended array of patterns from objects~(in Fig.~\ref{fig:adversarial_drop}\textcolor{linkcolor}{a}),
thus making precise predictions against corruption.

We have extensively validated the effectiveness of our proposed method on multiple benchmarks, including ModelNet-C~\cite{ren2022benchmarking} and the more challenging ScanObjectNN-C~\cite{wang2023adaptpoint}. The results significantly underscore improvements in robustness and establish state-of-the-art performance on these datasets. In addition, our algorithm demonstrates pronounced generalization capabilities in downstream tasks, as evidenced by its adeptness in shape segmentation under corruptions in ShapeNet-C~\cite{ren2022pointcloud}.
These findings collectively underscore the method's generality and effectiveness in enhancing the robustness of point cloud perception models.
Our contributions can be summarized as follows:
\vspace{-2mm}
\begin{itemize}
    \item We present a new framework, APCT, which effectively improves the resilience of point cloud model against various types of corruptions by leveraging adversarial mining strategy.
    \item We propose a novel adversarial dropout method that can stimulate the model to embrace broader valuable patterns by adversarially lifting dropout probability of specific patterns. 
    \item We demonstrate the effectiveness of APCT through extensive experiments on multiple point cloud benchmarks under diverse corruption scenarios.
\end{itemize}

\vspace{-4mm}

\section{Related Work}

\vspace{-2mm}
\subsection{Transformer for Point Cloud Recognition}
\vspace{-2mm}
Amidst the burgeoning epoch of vision transformers, seminal works such as Point Transformer~\cite{zhao2021point} and Point Cloud Transformer~\cite{guo2021pct} ventured into the realm of leveraging attention mechanisms tailored for point cloud semantics, thus pioneering this trajectory. The Point Cloud Transformer intrinsically integrates global attention across the entirety of the point cloud, mirroring certain challenges inherent to the Vision Transformer (ViT), notably the constraints of memory bandwidth and computational overhead. Contrarily, the Point Transformer accentuates local attention, distinctly focusing on each point and its proximate neighbors, effectively mitigating the aforementioned memory impediments. Subsequent to these, the Robust Point Cloud Classifier (RPC)~\cite{ren2022benchmarking} emerged, proffering a robust framework for point cloud classification. Infusing 3D representation, k-NN, frequency grouping, and self-attention, it posits itself as a formidable baseline in the annals of point cloud robustness inquiries.
While these methods demonstrate remarkable accuracy , their performance tends to deteriorate in face of real-world corruptions. In this manuscript, we introduce a novel transformer-based model specifically architected to robustly handle real-world corruptions.

\vspace{-2mm}
\subsection{Robust Learning Against Corruptions}
\vspace{-2mm}
In the realm of point cloud perception, deep learning has made significant strides~\cite{qi2017pointnet, qi2017pointnet++, ma2021rethinking, qianpointnext}. However, ensuring model robustness in the presence of corruptions remains a pivotal concern. Contemporary literature underscores the imperative of addressing point cloud corruptions~\cite{ren2022benchmarking,ren2022pointcloud,kim2021point,lee2021regularization}. Some methodologies have hinged on preprocessing, incorporating denoising or completion strategies~\cite{zhou2019dup}, to bolster a model's resilience against adversarial perturbations. Others have pursued augmentation pathways, spanning mix-based~\cite{lee2021regularization} and deformation-centric~\cite{kim2021point, ren2022benchmarking} techniques. The advent of advanced auto-augmentations~\cite{wang2023adaptpoint} has ushered in sample-adaptive data augmentations, targeting enhanced model robustness. Notwithstanding their merits, such interventions often mandate augmented computational overhead and may falter in scenarios of pronounced corruption.
Another line of research aims to improve the robustness of deep learning architectures themselves. 
PointASNL~\cite{yan2020pointasnl} integrates adaptive sampling coupled with local-nonlocal modules, striving for heightened robustness. Concurrently, Ren \textit{et al.}~\cite{ren2022pointcloud} unveiled an attention-based backbone, RPC, tailored to withstand corruption. Nevertheless, these methods only make improvements to the feature extractor, without concerning the nature of the corruption problem. This manuscript introduces to leverage an adversarial dropping strategy to discern pertinent features from corrupted samples.

\vspace{-3mm}
\section{Methodology}
\label{label:method}
\vspace{-3mm}

\begin{figure*}[t!]
  \centering
  \setlength{\abovecaptionskip}{2pt}  
  \setlength{\belowcaptionskip}{2pt}
  \includegraphics[width=0.95\textwidth]{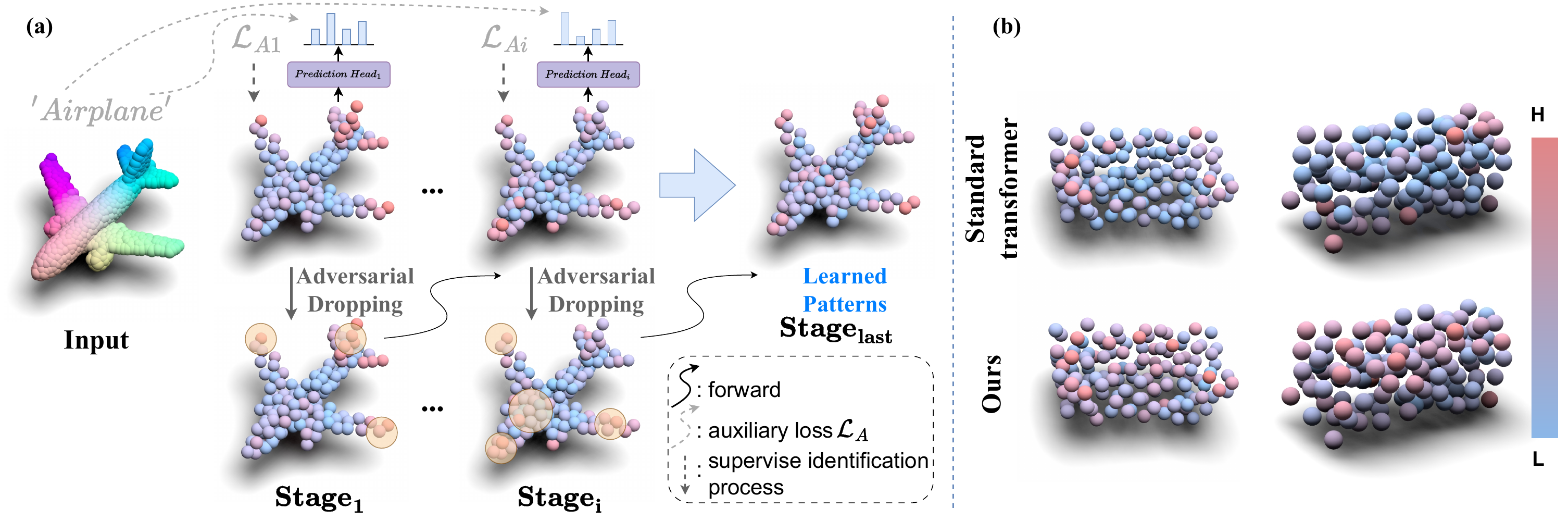}
  \caption{(a)~Process of progressive adversarial dropping. The first line means token learned in each stage.
  (b)~Visualization of token weights learned by the classifier. 
  Compared with standard transformer, ours has an advantage in mining  sample patterns.}
  \label{fig:adversarial_drop}
\vspace{-6mm}
\end{figure*}

We instead devise an adversarial analysis based framework (Fig.~\ref{fig:adversarial_drop}), which not only learns point recognition with current vital patterns, but more essentially, it automatically discovers and emphasizes the sub-important patterns, with the auxiliary supervised pattern miner.
Patterns learned in such strategy are expected to be more discriminative and robust, hence facilitating final recognition of point clouds under corruptions.
At each training iteration, our algorithm comprises of two phases. 
In \textbf{phase 1}, we perform adversarial mining over across all patterns, based on the token significance identification process. The propose is to search for the sub-optimal patterns that contribute less to the final recognition.  
In \textbf{phase 2}, we leverage deterministic dropping assignments and attribute elevated dropping rates to tokens pivotal for the final perception, as an  constraint to enable these sub-optimal tokens to play a increasing significant role in perceptual.
Engaging in this iterative adversarial procedure allows the network to progressively unearth and assimilate a comprehensive set of object patterns, as seen in Fig.~\ref{fig:adversarial_drop}\textcolor{linkcolor}{b}, culminating in refined predictions in the face of corruption.

\subsection{Overview Implementations}
\begin{figure*}[!t]
  \centering
  \setlength{\abovecaptionskip}{1pt}  
  \setlength{\belowcaptionskip}{-5pt}
  \includegraphics[width=0.98\textwidth]{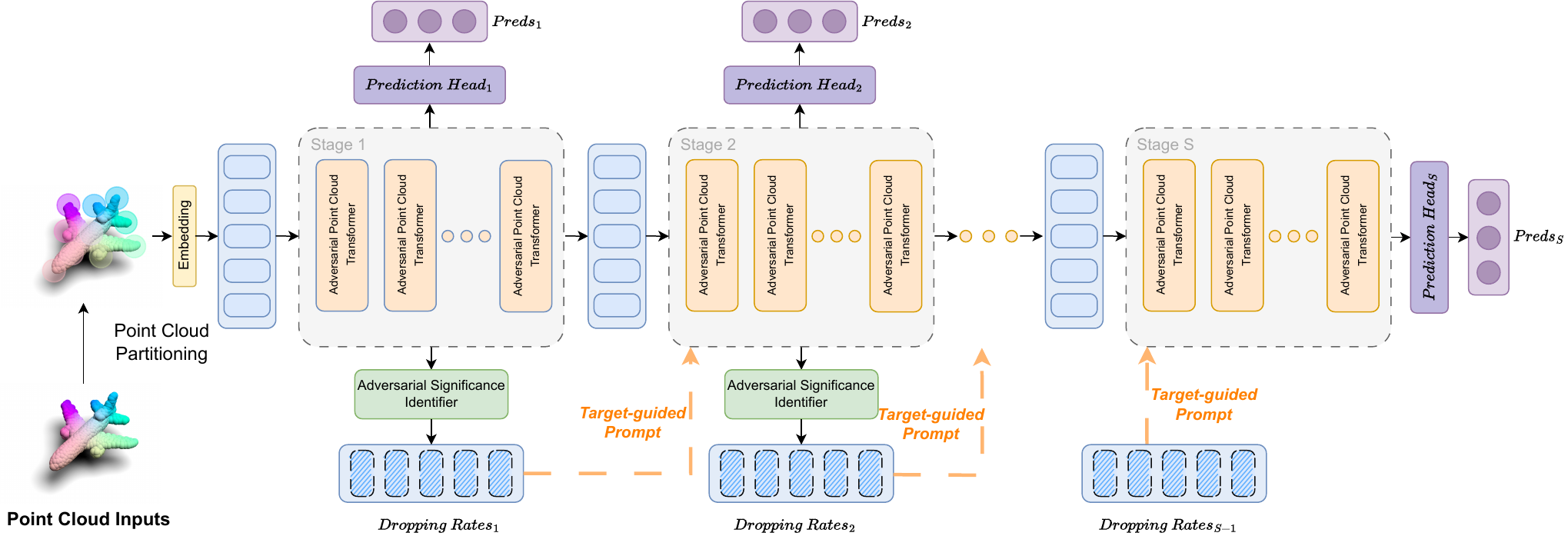}
  \caption{\textbf{Overall architecture} of our algorithm, composed of two key modules: {Adversarial Significance Identifier} and {Target-guided Promptor}. 
  The former evaluates token significance within the context of the global perception, with the help of dominant feature indexing process from an auxiliary supervising loss that can bolster the precision of the index selection, then producing dropping rate for tokens.
  Subsquently, {Target-guided Promptor} enhances key dropout probabilities influenced by rate above, driving the model to explore auxiliary segments for pivotal information.
  This mechanism mitigates the propensity of the model to overfit to localized patterns. 
  }
  \label{fig:pipeline}
\vspace{-4mm}
\end{figure*}

\vspace{-2mm}
The overall implementations of Adversarial Point Cloud Transformer is shown in Fig.~\ref{fig:pipeline}. Given an input set of point cloud $\mathcal{P} \in \mathbb{R}^{N \times 3}$,
we first partition the input point cloud into an assembly of $n$ discrete patches, and subsequently, generating pertinent tokens with $C$-dimensional features, $T = \left\{ t_j | j = 1,...,n \right \} \in \mathbb{R}^{n\times C}$.
Taking these tokens as input for the following stages, the network initially employs the {Adversarial Significance Identifier} module to formulate a per-token dropping rate, assigning higher rates to tokens that are critical for classification. Subsequently, the {Target-guided Promptor} selectively eliminates key vectors based on the established rates. 
Through introducing such an adversarial progressive dropping scheme, the network gathers a diverse set of perception cues and assimilates various patterns from an expansive region of the underlying object structure, thus enhancing classifier resilience against potential corruptions.

The proposed architecture is segmented into three distinct stages, each consisting of multiple blocks. Within stages, the depth of self-attention blocks is conventionally configured to [4, 4, 4]. Each stage is further equipped with a uniformly applied {Adversarial Significance Identifier} and an associated complementary supervision head. The default adversarial dropping ratio, is uniformly set to [0.2, 0.2, 0.2] across all stages.
Each block within stages independently incorporates a {Target-guided Promptor}.

\subsection{Adversarial Significance Identifier}
\begin{figure*}[t!]
\begin{minipage}[t]{0.46\textwidth}
  \setlength{\abovecaptionskip}{-3pt}  
  \setlength{\belowcaptionskip}{-5pt}
\includegraphics[width=\textwidth]{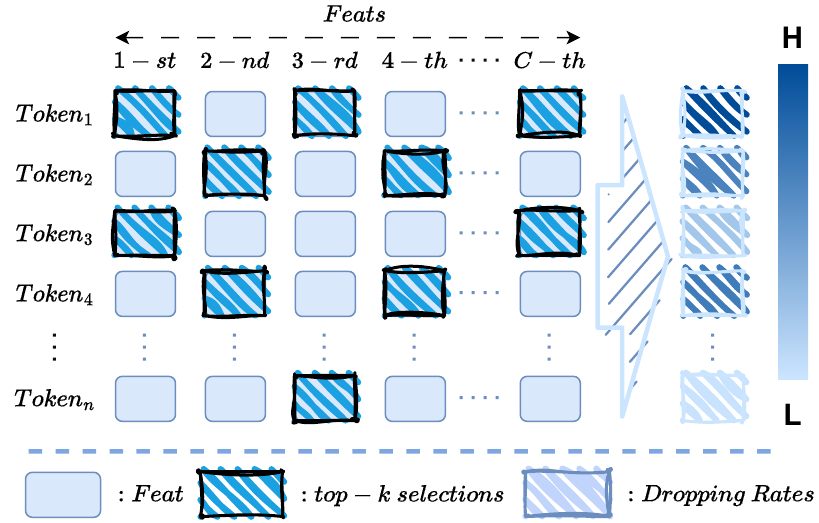}
\figcaption{ \textbf{Workflow of Adversarial Significance Identifier}. Squares with \textcolor{lightblue}{light}/\textcolor{darkblue}{dark} blue means \textcolor{lightblue}{low}/\textcolor{darkblue}{high} values, respectively. } 
\label{fig:Index Selection}
\end{minipage}
\hspace{0.02in}
\begin{minipage}[t]{0.50\textwidth}
  \setlength{\abovecaptionskip}{-5pt}  
  \setlength{\belowcaptionskip}{-5pt}
\includegraphics[width=\textwidth]{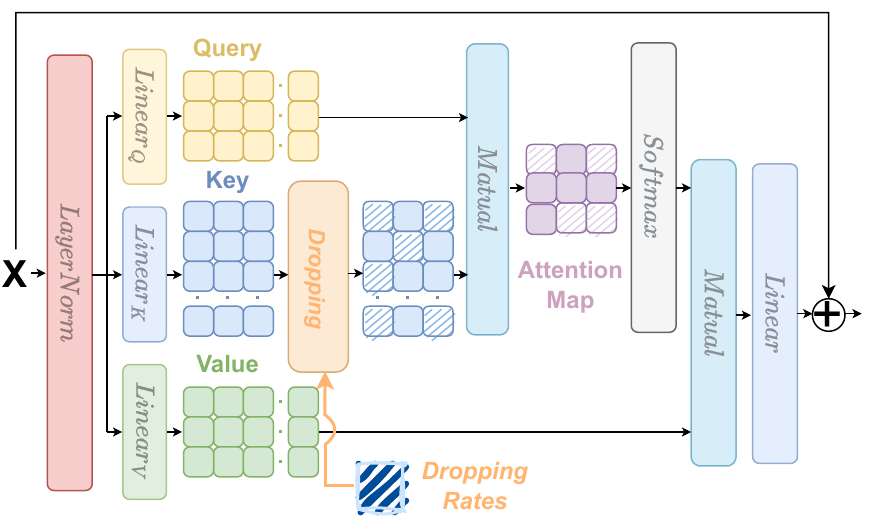}
\figcaption{\textbf{The proposed Target-guided Prompting} mechanism that increases the dropout probability, based on the per-token dropping rate. } 
\label{fig:dropkey_attention}
\end{minipage}
\vspace{-4mm}
\end{figure*}

\vspace{-2mm}
\noindent This module operates by meticulously processing the input tokens extracted from point clouds.
It evaluates and distinguishs the relative significance of each token and applies constraints particularly to those tokens identified as most crucial in the context of point representation learning, 
thereby efficiently unveiling underlying data structures through progressively mining patterns that sub-significant yet representative and insightful.
Utilizing an auxiliary supervisory process, it integrates global contextual analysis and discerns significance of all input tokens $T$; thus identifying the focal tokens, finally generating per-token dropout rate. 
To this end, it sorts tokens based on the feature channel responses via index number in auxiliary supervision process, then it establishes the mapping between tokens and the dropout rate, finally predict per-token dropout rate $M = \left\{ m_j | j = 1,...,n \right \} \in \mathbb{R}^{n}$.
The overall workflow is shown in Fig.~\ref{fig:Index Selection}.

\noindent\textbf{Focal tokens identification.}
In the context of point cloud analysis, let $T = \left\{ t_j | j = 1,...,n \right \} \in \mathbb{R}^{n\times C}$ represent a set comprising point cloud tokens. Initially, an essential step entails the organization of token contributions to final perception based on their associated feature channel responses, through the auxiliary supervisory process. Subsequent to this arrangement, we proceed with the identification of the most prominent tokens pertaining to each feature channel, resulting in the derivation of their respective index bank denoted as $F_{\text{topk}}= \left\{ f_{\text{topk}}^{m} | m = 1,...,D \right \} \in \mathbb{R}^{k\times C}$, where $k$ signifies the number of tokens retained per channel feature. This process can be succinctly expressed as:
\vspace{-0.5mm}
\begin{equation}
\label{eq:topk}
    F_{topk} = {TopK}(T),
\end{equation}
where $f_{\text{topk}}^{m} \in \mathbb{R}^{k}$ enumerates values within the range $(1, N)$, indicating the token associated with the feature response under consideration.
Upon the construction of the index repository $F_{topk}$, we perform a comprehensive aggregation of the elements within it. This process is pivotal for quantifying the occurrence frequency of each token in the ultimate perception of the model. This assists in delineating the tokens of substantial relevance. 
The mathematical representation is as follows:
\vspace{-0.5mm}
\begin{equation}
    M = \psi (F_{topk}),
\end{equation}
where $\psi$ symbolizes the function employed to compute the token significance from $ F_{topk} $ in accordance with predefined indices.
Consequently, the matrix $M = \left\{ m_j | j = 1,...,n \right\} \in \mathbb{R}^{n}$ emerges as the importance matrix for tokens. Each element within this matrix, denoted by $M_j \in \mathbb{R}^{1}$, serves to quantify the representation significance of the $j-th$ token within the overall perception framework.

\noindent\textbf{Supervisory token identification process.}
Current algorithms that optimize the network under the guidance of readily high-level labels at final layer is insufficient for guiding the selection of drop matrices in the intermediate stages of the network. 
This limitation may lead to the phenomena that constructed matrices do not accurately reflect the contributions of tokens, potentially compromising the efficiency of algorithm. Thus, the major question arises: \textit{how can we optimize the selection of drop matrices to accurately represent each token's contribution?}

To rectify this, our framework innovates with an adversarial approach embedded within each stage, which cornerstone is the integration of specialized auxiliary heads, strategically positioned within the network. Each auxiliary head is assigned the critical task of computing a unique and stage-specific loss, 
it identifies the patterns or tokens that the current stage predominantly focuses on and then intentionally drops these identified tokens. This process forces the network to shift its attention, thereby encouraging it to mine features from other, less emphasized regions.

The operation within these auxiliary heads begins with a dominant feature index operation applied to the tokens, which aligns with the identification methodology detailed in Eq.~\ref{eq:topk}. This ensures that the drop matrices generated are a true reflection of tokens significance at every specific network stage.

Implementing these adversarial-focused auxiliary heads represents a significant leap in the optimization process. By introducing a targeted, stage-specific supervision issue at each stage, they offer a refined assessment of token contributions, a stark contrast to the broader, less specific results from the network's final layer. This precision in evaluation is pivotal to the nuanced understanding and optimization of each stage within the network, aligning token signification closely with the specific  objectives of each stage in the network.
The adversarial approach facilitates an augmented constraint term and plays an instrumental role in the selection of drop matrices at each stage, contributing substantially to the optimization process,  
encapsulated as follows:
\vspace{-2mm}
\begin{equation}
\label{eq:hier loss}
    \mathcal{L}_{A} = \frac{1}{N}\sum_{i=1}^{N} \left \| {y^*}_i - {\hat{y}}_i \right \|_2,
`\end{equation}
where ${\hat{y}}$ and ${y^*}$ denote the predicted and ground-truth labels, respectively, for the input point cloud $\mathcal{P} \in \mathbb{R}^{N\times 3}$, with $\left \| \cdot  \right \|_2$  indicating the $L_2$ norm. 
This constraint term, derived from the cumulative loss computed by all the auxiliary heads, ensures a comprehensive and layered supervision across the network. It allows the Adversarial Significance Identifier to discern the contribution of each token more accurately, thereby generating more targeted and effective supervisory signals for the token dropping process. This, in turn, enhances the overall efficacy and precision of the network learning. 

\noindent\textbf{Derivation of per-token dropout rate.}
The objective of $\bm{\Phi}(\cdot)$ below is to refine the raw frequency distribution of tokens, as encapsulated within the matrix $M$, transforming them into precise and actionable probability distribution. This transformation is critical for the effective application of dropout rates to tokens in the network. The specific functional representation is detailed below:
\vspace{-2mm}
\begin{equation}
\label{eq: mapping function}
\bm{\Phi}(m_j, \gamma, \alpha, \beta) = 
\begin{cases}
\alpha & \text{if } \gamma  \frac{n \cdot m_j  } {\sum_{j=1}^{n} m_j} < \alpha \\
\gamma  \frac{n \cdot m_j  } {\sum_{j=1}^{n} m_j} & \text{if } \alpha \leq \gamma  \frac{n \cdot m_j  } {\sum_{j=1}^{n} m_j} \leq \beta \\
\beta & \text{if } \gamma  \frac{n \cdot m_j  } {\sum_{j=1}^{n} m_j} > \beta \\
\end{cases}
\end{equation}
Herein, the parameter $\gamma$ represents the mapping ratio, a pivotal factor whose impact and intricacies are thoroughly examined in a later ablation study. The variable $n$ denotes the total number of tokens under consideration. The boundary parameters, $\alpha$ and $\beta$, are pre-set to the values of $0.05$ and $0.95$, respectively. These parameters play a vital role in defining the range and sensitivity of the transformation function.
The probability distribution thus generated by $\bm{\Phi}(\cdot)$ is instrumental in dictating the dropout strategy in the following {Target-guided Promptor}. This structured approach ensures that the dropout process is both targeted and efficient, enhancing the overall performance.

\subsection{Target-guided Promptor}
\vspace{-2mm}
\noindent In traditional self-attention mechanism in transformers, the use of vanilla dropout techniques is prevalent, where each node in attention matrix is assigned a uniform stochastic discard probability. While this strategy aids in reducing overfitting and enhancing the generalization, it inherently overlooks the non-uniform characteristics of nodes in the matrix. The indiscriminate application of random dropout across all nodes fails to effectively penalize those with significantly higher attention scores; this oversight may result in a residual risk of overfitting to specific localized patterns.
To address this shortcoming, our proposed methodology diverges from the conventional approach by independently targeting each key within the self-attention mechanism. As detailed in Fig.~\ref{fig:dropkey_attention}, this focused strategy efficiently penalizes keys with pronounced attention scores, thereby directly addressing the issue of overfitting to specific local patterns.

During each training epoch, this module adaptively masks a specified fraction of keys in the input key map using the matrix $M$. Notably, for every distinct query, a unique masked key map is synthesized, as opposed to utilizing a shared masked key map for the entire set of query vectors.
Given token features defined by $T = \left\{ t_j | j = 1,...,n \right \} \in \mathbb{R}^{n\times C}$ accompanied by coordinates $\mathcal{D}$, we derive the representations for query ($Q$), key ($K$) and value ($V$) as following:
\vspace{-1mm}
\begin{equation}
     Q, K, V = H {W}^{Q},\ H {W}^{K},\ H {W}^{V},
\end{equation}
where $ {W}^{Q}, {W}^{K}, {W}^{V} \in \mathbb{R}^{C \times C}$ are the learnable linear projections.

\noindent\textbf{Generating dropout target.}
We aim to generate a vector $M'$ whose elements have probability of being negative infinity, based on the value in $M$ above.
Based on the probability values encapsulated within matrix $M$, we synthesize matrix $M'$. Each element within $M'\in \mathbb{R}^{n \times C}$ is assigned a value of negative infinity with a probability determined by the corresponding value in $M$:
\vspace{-1mm}
\begin{equation}
    m'_{i} = \begin{cases}
0 &  with~probability~ 1 - m_i \\
-inf & with~probability~ m_i \\
\end{cases}
\end{equation}
This resultant vector is then expanded across feature channels.

\noindent\textbf{Dropping key process.}
In preceding step, we derived the dropout rates $M'$. This is subsequently employed to drop elements from the key vector $K$. Specifically, 
by adding the vector $M'$, which contains negative infinity values, to $K$,  the network disregards designated positions, thus achieving a targeted dropout implementation.
The mechanism is defined as follows:
\vspace{-1mm}
\begin{equation}
     K' = K + M',
\end{equation}
The modified token features $T' = \left \{ t_i' | i = 1,...,n \right \} \in \mathbb{R}^{n \times C}$ are derived by such formulation:
\vspace{-1mm}
\begin{equation}
    T' = MHA(Q, K', V, PE(\mathcal{D})),
\end{equation}
where $MHA$ symbolizes the multi-head attention mechanism as delineated in~\cite{vaswani2017attention}, while $PE(\cdot)$  is indicative of the position embedding.

\subsection{Overall Learning Objectives. }
\vspace{-2mm}
We present the adversarial analysis based framework that leverages an
{Adversarial Significance Identifier} along with a {Target-guided Promptor}. The loss function comprises of the cross-entropy loss $\mathcal{L}_{C}$ and the auxiliary token identification loss $\mathcal{L}_A$, combined to form the complete loss function:
\vspace{-2mm}
\begin{equation}
    \mathcal{L} = \mathcal{L}_{C} + \lambda \sum_{s=1}^{S-1} \mathcal{L}_{A},
\end{equation}
where $\lambda$ is a balancing weight, set as $1$. $S$ denotes the total stages in the network.

\section{Experiments}
\label{label:experiments}
\vspace{-2mm}
We first report our 3D robust classification results on 
synthetically generated and real-scanned datasets in \S\ref{para: results on modelnet-c} and \S\ref{para: results on scanobjectnn-c}, respectively. Subsequently, we assess our methodology efficacy in the robust 3D segmentation 
in \S\ref{para: results on Shapenet-c}.
In \S\ref{para: ablation study}, we provide ablative analyses on 
our core algorithm design.

\subsection{Results on ModelNet-C}
\label{para: results on modelnet-c}
\vspace{-2mm}
\noindent\textbf{Dataset.}
We train models on the clean ModelNet40~\cite{modelnet40} dataset and evaluate them on the ModelNet-C~\cite{ren2022benchmarking} corruption test suite,
which includes seven types of corruptions, “Jitter”, “Drop Global/Local”, “Add Global/Local”, “Scale” and “Rotate”, with five levels of severity . 
We take mean corruption error metric (\textcolor{citegrey}{mCE, $\%, \downarrow$}) as the main evaluation matrix.
More details are in the supplementary materials.

\begin{table*}[t!]
    \centering
    \setlength{\tabcolsep}{6.0pt}
    \renewcommand\arraystretch{0.85}
    \setlength{\abovecaptionskip}{-2pt}  
    \setlength{\belowcaptionskip}{2pt}
    \caption{Classification results on the ModelNet-C dataset, mCE($\%, \downarrow$) is reported, the best performance is \textbf{bold}.  $\dagger$ denotes method designed specifically against corruption.}
    \label{tab:modelnet40-c, mCE}
    \footnotesize
    \begin{tabular}{l|c|cccccccc}
        \toprule 
        Method  & ~mCE~  & Scale & Jitter & Drop-G & Drop-L & Add-G & Add-L & Rotate \\ 
        \midrule
        DGCNN\pub{TOG2019}~\cite{wang2019dynamic}  & 100.0  & 100.0  & 100.0  & 100.0  & 100.0  & 100.0  & 100.0  & 100.0  \\
        PointNet\pub{CVPR2017}~\cite{qi2017pointnet} & 142.2  & 126.6  & {64.2}  & 50.0  & 107.2  & 298.0  & 159.3  & 190.2  \\ 
        PointNet++\pub{NIPS2017}~\cite{qi2017pointnet++}  & 107.2  & 87.2  & 117.7  & 64.1  & 180.2  & 61.4  & 99.3  & 140.5  \\ 
        GDANet\pub{AAAI2021}~\cite{xu2021learning}  & 89.2  & {83.0}  & 83.9  & 79.4  & 89.4  & 87.1  & 103.6  & 98.1  \\ 
        PCT\pub{CVM2021}~\cite{guo2021pct} & 92.5  & 87.2  & 87.0  & 52.8  & 100.0  & 78.0  & 138.5  & 104.2  \\ 
        RPC$^{\dagger}$\pub{ICML2022}~\cite{ren2022benchmarking} & 86.3  & 84.0  & 89.2  & 49.2  & {79.7}  & 92.9  & 101.1  & 107.9  \\ 
        PointNeXt\pub{NIPS2022}~\cite{qianpointnext} & 85.6  & 90.4  & 129.7  & 84.7  & 95.7  & {25.1}  & {27.6}  & 146.0 \\
        PointM2AE\pub{NIPS2022}~\cite{zhang2022pointm2ae} & 83.9 &	95.7 &	159.2 &	53.2 &	83.6 &	30.2 &	30.5 &	134.9 \\
        PointGPT\pub{NIPS2023}~\cite{chen2024pointgpt} & 83.4 & 108.5 & 	102.5 & 	58.5 & 	98.1	 & 29.2 & 	32.7 & 	154.4\\
        \midrule
        \rowcolor{RowColor} APCT (Ours) & \textbf{72.2} & 94.7 & 88.3 & {46.8} & 85.0 & 28.5 & 29.8 & 132.6 \\
        \fontsize{7.5pt}{1em}\selectfont{\color{updatagrey}{\textit{~{vs. prev. SoTA}}}} & \fontsize{7.5pt}{1em}\selectfont{\color{updatagrey}{\textsl{$\downarrow$\textbf{11.2}}}}  & \multicolumn{7}{c}{\ \ \ - \ \ \ } \\
        \bottomrule
    \end{tabular}
\vspace{-6mm}
\end{table*}

\noindent\textbf{Main Results.}
The comparative performance of various methodologies in terms of their resilience to corruption is systematically encapsulated in Tab.~\ref{tab:modelnet40-c, mCE}.
The results unequivocally demonstrate that our proposed method exhibits superior performance, registering an exemplary state-of-the-art mCE score of $\textbf{72.2\%}$. This outstanding performance is notably achieved through straightforward network architectural modifications paired with our unique adversarial approach, eschewing the need for intricate training scheme alterations or complex feature extraction. In direct comparison with the sota method PointGPT~\cite{chen2024pointgpt}, our method exhibits a marked enhancement of $11.2\%$ mCE reduction. 
Moreover, APCT consistently attains high mCE scores across categories: $46.8\%$, $85.0\%$, $28.5\%$, and $29.8\%$ for drop-global, drop-local, add-global, and add-local, respectively. These metrics are either at the pinnacle or are approaching the current best results in each respective subcategory. 
Such results underscore the efficacy of APCT in maintaining robustness against a wide spectrum of corruptions, encompassing both global and local perturbations
When compared with RPC~\cite{ren2022benchmarking}, an architecture tailored for enhanced robustness against corruption, APCT manifests a striking improvement in mCE for add-global, surpassing $60\%$. This underlines APCT proficiency in discerning subtle global geometric intricacies within the point cloud, thus facilitating the extraction of pivotal features.

\begin{table}[!t]
    \centering
    \setlength{\tabcolsep}{6.0pt}
    \renewcommand\arraystretch{0.85}
    \setlength{\abovecaptionskip}{-2pt}  
    \setlength{\belowcaptionskip}{2pt}
    \caption{Classification results on the ScanObjectNN-C dataset, mCE($\%, \downarrow$) is reported.  $\dagger$ denotes method designed specifically against corruption.}
    \label{tab:ScanObjectNN-c, mCE}
    \footnotesize
    \begin{tabular}{l|c|cccccccc}
        \toprule 
        Method & ~mCE~ & Scale & Jitter & Drop-G & Drop-L & Add-G & Add-L & Rotate \\ 
        \midrule
        DGCNN\pub{TOG2019}~\cite{wang2019dynamic} & 100.0  & 100.0  & 100.0  & 100.0  & 100.0  & 100.0  & 100.0  & 100.0  \\ 
        PointNet\pub{CVPR2017}~\cite{qi2017pointnet} & 132.8  & 141.2  & {88.4}  & 78.3  & 125.3  & 139.2  & 200.0  & 156.9  \\ 
        PointNet++\pub{NIPS2017}~\cite{qi2017pointnet++} & 96.9  & 89.7  & 110.3  & 55.0  & 127.7  & 94.7  & 90.5  & 110.7  \\ 
        RPC$^{\dagger}$\pub{ICML2022}~\cite{ren2022benchmarking} & 132.6  & 131.7  & 107.3  & 145.5  & 130.5  & 114.2  & 158.7  & 140.2  \\
        PointNeXt\pub{NIPS2022}~\cite{qianpointnext} & 92.1  & {80.3}  & 107.9  & 80.7  & 94.2  & 94.4  & 87.5  & {99.5} \\
        PointM2AE\pub{NIPS2022}~\cite{zhang2022pointm2ae} & 78.2 & 	68.5 & 	119.3 &  55.3 & 	83.2 & 	35.7 & 	71.4 &  114.2 \\
        PointGPT\pub{NIPS2023}~\cite{chen2024pointgpt} & 84.7	& 	119.9 & 104.0 & 	  58.5 & 		76.9 & 		36.1 & 		72.2 & 	 125.1\\
        
        \midrule
        \rowcolor{RowColor} APCT (Ours) & \textbf{74.2} & 104.5 & 98.9 & {48.9} & {67.0} & {30.0} & {63.0} & 107.1 \\
        \fontsize{7.5pt}{1em}\selectfont{\color{updatagrey}{\textit{~{vs. prev. SoTA}}}} & \fontsize{7.5pt}{1em}\selectfont{\color{updatagrey}{\textsl{$\downarrow$\textbf{4.0}}}}  & \multicolumn{7}{c}{\ \ \ - \ \ \ } \\
        \bottomrule
    \end{tabular}
\vspace{-6mm}
\end{table}

\subsection{Results on ScanObjectNN-C}
\label{para: results on scanobjectnn-c}
\vspace{-2mm}
\noindent\textbf{Dataset.}
ScanObjectNN consists of objects acquired from real-world scans, offering a more authentic evaluation of recongnition, particularly when compared with CAD datasets.
ScanObjectNN-C is the corresponding corruption test suite of the hardest version of ScanObjectNN.

\noindent\textbf{Main Results.}
Tab.~\ref{tab:ScanObjectNN-c, mCE} summarizes the comparison results on ScanObjectNN-C, showing our algorithm works well on real-world challenging corrupted point clouds. In particular, our algorithm achieves impressive sota result of $\textbf{74.2\%}$ mCE. 
In line with the other results, our method distinctly surpasses them, delivering $\textbf{4.0\%}$ and $\textbf{7.5\%}$ mCE reduction for PointM2AE~\cite{zhang2022pointm2ae} and PointGPT~\cite{chen2024pointgpt}, respectively.
This confirms our algorithm is applicable in real-world corruptions.
\noindent Notably, our method obtains consistent high performance of nearly all categories. it achieves $48.9\%$, $67.0\%$, $30.0\%$, and $63.0\%$ mCE over drop-global, drop-local, add-global, and add-local, respectively.
These results are all the sota results in each subcategory, illustrating that our method performs well in the face of point adding and removing corruption, both globally and locally.

\subsection{Results on Shapenet-C}
\label{para: results on Shapenet-c}
\vspace{-2mm}
\noindent\textbf{Dataset.}
Our method can generalize well to down-stream tasks, including part segmentation. To validate this, we conduct a part segmentation experiment on the ShapeNet-C dataset~\cite{ren2022pointcloud}.

\noindent\textbf{Main Results.}
Tab.~\ref{tab:results_shapenet-c} showcases the performance of our method, achieving a mCE score of $\textbf{81.4\%}$ and surpassing the previous sota GDANet and PointMAE by an impressive margin of $10.9\%$ and $11.3\%$, respectively. It is worth highlighting that PointMAE operates as a scheme of masked autoencoders for point cloud self-supervised learning, which is pretrained on large set of data. 
This underscores the substantial enhancement brought about by our technique in improving the model robustness 

\begin{table}[!h]
    \centering
    \setlength{\tabcolsep}{4.3pt}
    \renewcommand\arraystretch{0.85}
    \setlength{\abovecaptionskip}{-2pt}  
    \setlength{\belowcaptionskip}{2pt}
    \caption{Segmentation results in terms of mCE ($\%, \downarrow$) on ShapeNet-C dataset.} 
    \label{tab:results_shapenet-c}
    \begin{tabular}{l|c|ccccccc}
        \toprule 
        Method & mCE & Scale & Jitter & Drop-G & Drop-L & Add-G & Add-L & Rotate \\ 
        \midrule
        DGCNN\pub{TOG2019}~\cite{wang2019dynamic} & 100.0  & 100.0  & 100.0  & 100.0  & 100.0  & 100.0  & 100.0  & 100.0  \\
        PointNet\pub{CVPR2017}~\cite{qi2017pointnet} & 117.8  & 108.2  & 105.0  & 98.3  & 113.2  & 138.6  & 117.3  & 143.8  \\
        PointNet++\pub{NIPS2017}~\cite{qi2017pointnet++} & 111.2  & 95.0  & 108.1  & 85.6  & 198.3  & 88.6  & 108.3  & 94.7  \\
        PAConv\pub{CVPR2021}~\cite{xu2021paconv} & 92.7  & 92.7  & 107.2  & 92.5  & 92.7  & 74.3  & 94.8  & 94.8  \\ 
        GDANet\pub{AAAI2021}~\cite{xu2021learning} & 92.3  & 92.2  & \textbf{101.2}  & 94.2  & 94.6  & 71.2  & 95.7  & 96.9  \\ 
        PT\pub{ICCV2021}~\cite{zhao2021point} & 104.9  & 107.6  & 107.2  & 103.2  & 108.1  & 111.2  & 106.6  & 90.7  \\ 
        PointMLP\pub{ICLR2022}~\cite{ma2021rethinking} & 97.7  & 96.5  & 113.2  & 88.7  & 99.1  & 92.9  & 106.1  & \textbf{87.6}\\
        Point-BERT\pub{CVPR2022}~\cite{yu2022point_pointbert} & 103.3   & 93.8 & 109.8 & 87.3 & 92.7 & 117.0 & 119.9 & 102.5  \\
        Point-MAE\pub{ECCV2022}~\cite{pang2022masked} & 92.7  & \textbf{90.8}  & 103.5  & \textbf{85.2}  & \textbf{88.2}  & 77.6  & 103.1  & 100.3  \\
        \midrule
        \rowcolor{RowColor} APCT (Ours) & \textbf{81.4}	& 93.9	& 109.9 & 86.6 & 93.4 &	\textbf{40.1} & \textbf{41.6} & 103.8 \\
        \fontsize{7.5pt}{1em}\selectfont{\color{updatagrey}{\textit{~{vs. prev. SoTA}}}} & \fontsize{7.5pt}{1em}\selectfont{\color{updatagrey}{\textsl{$\downarrow$\textbf{10.9}}}}  & \multicolumn{7}{c}{\ \ \ - \ \ \ } \\
        \bottomrule
    \end{tabular}
\vspace{-5mm}
\end{table}

\subsection{Ablation Study}
\label{para: ablation study}
\vspace{-2mm}
To validate the efficacy of our core algorithm designs and parameter settings, we conduct a series of ablative studies on ModelNet-C~\cite{ren2022benchmarking}.

\begin{figure*}[t!]
\begin{minipage}[t!]{0.32\linewidth}
\centering
\footnotesize
\setlength{\tabcolsep}{7pt}
\renewcommand\arraystretch{0.9}
\setlength{\abovecaptionskip}{0pt}  
\setlength{\belowcaptionskip}{0pt}
\tabcaption{Ablation of Main Components in our method.}
\label{tab: ablation: dropping strategy}
    \begin{tabular}{cccc}
    \toprule
    \multirow{2.5}*{\textsl{Drop }}  &  \multirow{2.5}*{$\mathcal{L}_{A}$ } &  \multicolumn{2}{c}{\ \ \ Robust \ Cls.\ \ \ }   \\
    \cmidrule(lr){3-4}
     & & \scriptsize{MN-C}& \scriptsize{SCNN-C}\\
    \midrule
    \xmark & \xmark & 76.2 & 78.8\\
    \xmark & \cmark & 74.3 & 76.5\\
    \cmark & \xmark & 77.8 & 80.5\\
    \cmark & \cmark & \textbf{72.2} & \textbf{74.2}\\
    \bottomrule
    \end{tabular}
\end{minipage}
\qquad
\begin{minipage}[t!]{0.26\linewidth}
\centering
\footnotesize
\setlength{\tabcolsep}{7pt}
\renewcommand\arraystretch{0.9}
\setlength{\abovecaptionskip}{0pt}  
\setlength{\belowcaptionskip}{0pt}
 \tabcaption{Ablation of mapping ratio $\gamma$.}  
\label{tab: ablation: mapping ratio}
    \begin{tabular}{cc}
     \toprule
    $\gamma$ & mCE  \\
    \midrule
    ~ [0.1, 0.1, 0.1] & 78.1 \\
    ~ [0.2, 0.2, 0.2] & \textbf{72.2} \\
    ~ [0.3, 0.3, 0.3] & 76.5 \\
    \midrule
    ~ [0.1, 0.2, 0.3] & 77.4 \\
    ~ [0.3, 0.2, 0.1] & 76.9 \\
    \bottomrule
    \end{tabular}
\end{minipage}
\qquad
\begin{minipage}[t!]{0.31\linewidth}
\centering
\setlength{\abovecaptionskip}{0pt}  
\setlength{\belowcaptionskip}{0pt}
\figcaption{Ablation of selection number k} 
\label{tab: ablation: k number}
\includegraphics[width=\textwidth]{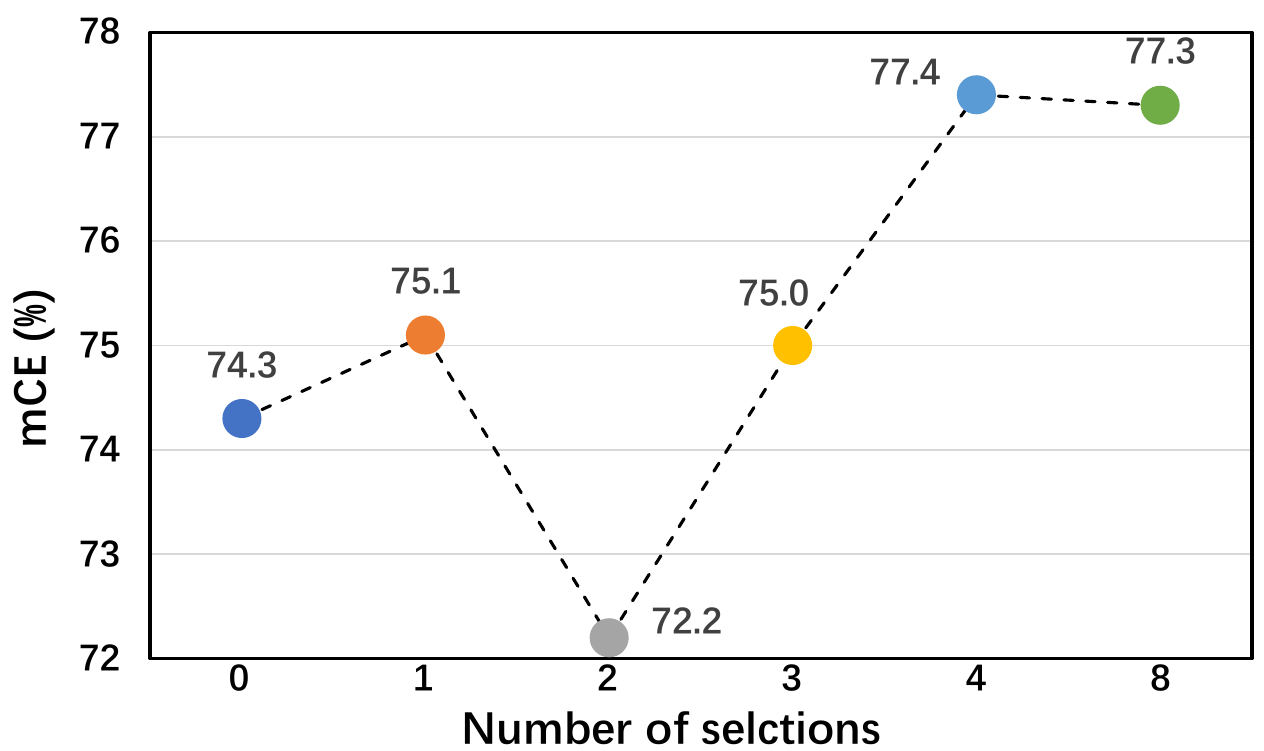}
\end{minipage}
\vspace{-6mm}
\end{figure*}

\noindent\textbf{Effect of adversarial dropping strategy.}
To ascertain the impact of our core idea of adversarial dropping, we conducted an ablation study by removing the adversarial dropping process, along with the collaborative supervising identification process.
As shown in Tab.~\ref{tab: ablation: dropping strategy}, the baseline model, trained in the standard strategy, gains $76.2\%$ and $78.8\%$ mCE, on ModelNet-C and ScanObjectNN-C, respectively. Additionally considering the dropping strategy and supervising identification process $\mathcal{L}_{A}$(Eq.~\ref{eq:hier loss}) can lead to $4.0\%$ and $4.6\%$ mCE reduction, respectively. However, adding dropping strategy without ancillary constraints from $\mathcal{L}_{A}$ will lead to mCE increasing. These results verify that combining these two training objectives can yield the best results, indicating that mining data structures from subsidiary component can benefit detailed analysis of point cloud

\begin{figure*}[tt!]
\begin{minipage}[t]{0.34\textwidth}
  \setlength{\abovecaptionskip}{2pt}  
  \setlength{\belowcaptionskip}{-5pt}
\includegraphics[width=\textwidth]{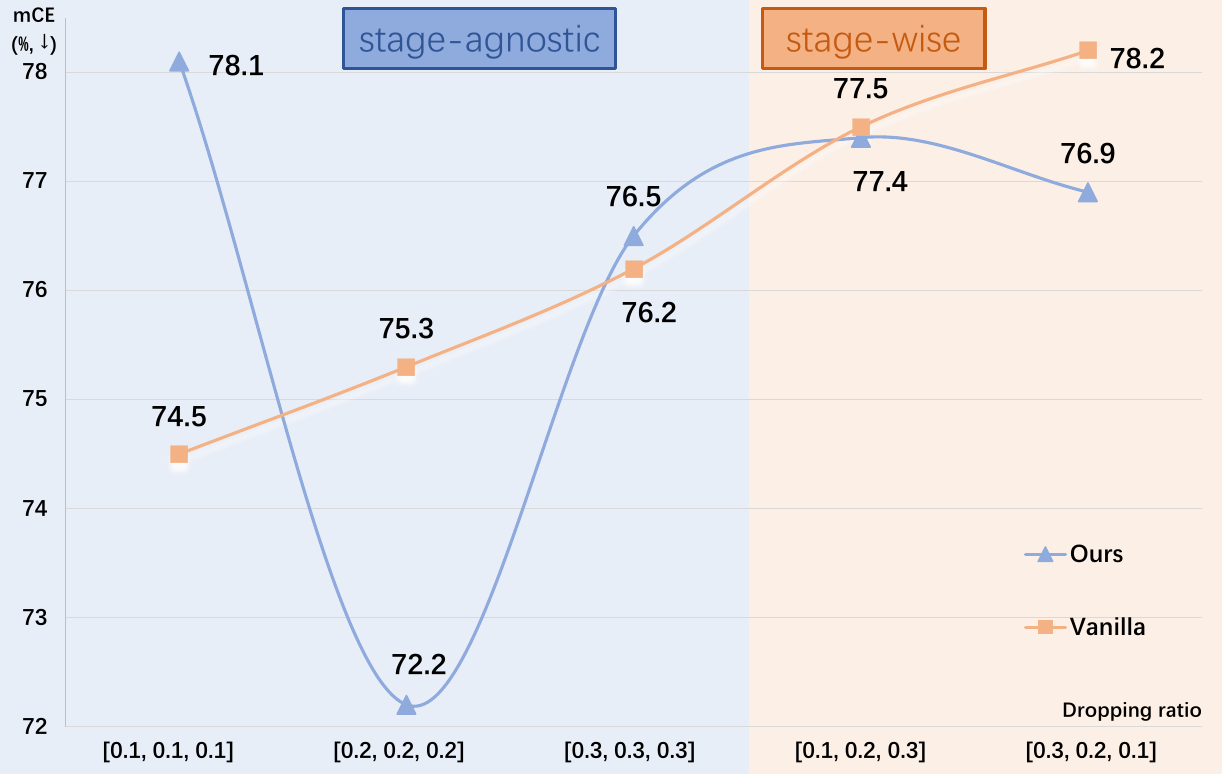}
\figcaption{\textbf{Comparative Curves} of our algorithm and vanilla dropout strategy.} 
\label{fig:comparison curves of apct and vanilla dropout}
\end{minipage}
\hspace{0.04in}
  \setlength{\abovecaptionskip}{2pt}  
  \setlength{\belowcaptionskip}{-5pt}
\begin{minipage}[t]{0.31\textwidth}
\includegraphics[width=\textwidth]{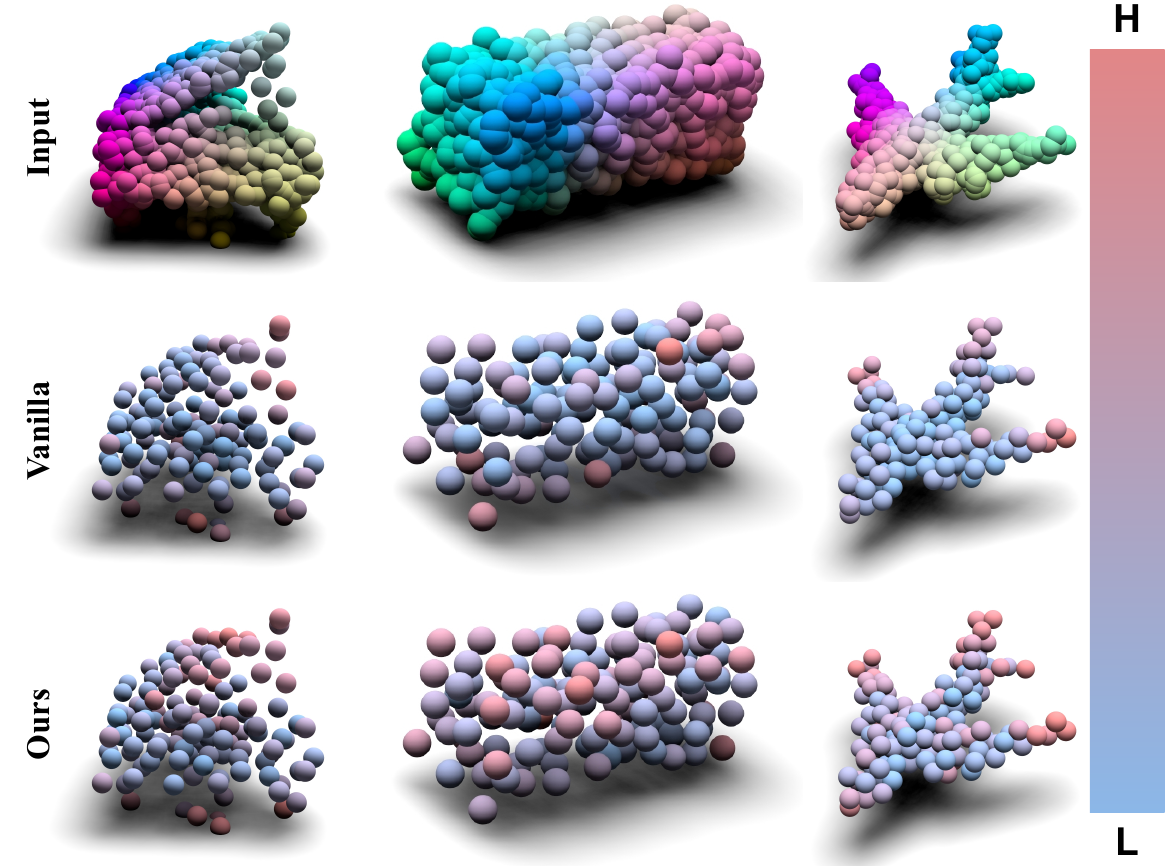}
\figcaption{Visualization of patterns learned by the classifier on ModelNet-C.}
\label{fig:fullcomparison vis weights}
\end{minipage}
\hspace{0.04in}
  \setlength{\abovecaptionskip}{2pt}  
  \setlength{\belowcaptionskip}{-5pt}
\begin{minipage}[t]{0.26\textwidth}
\includegraphics[width=\textwidth]{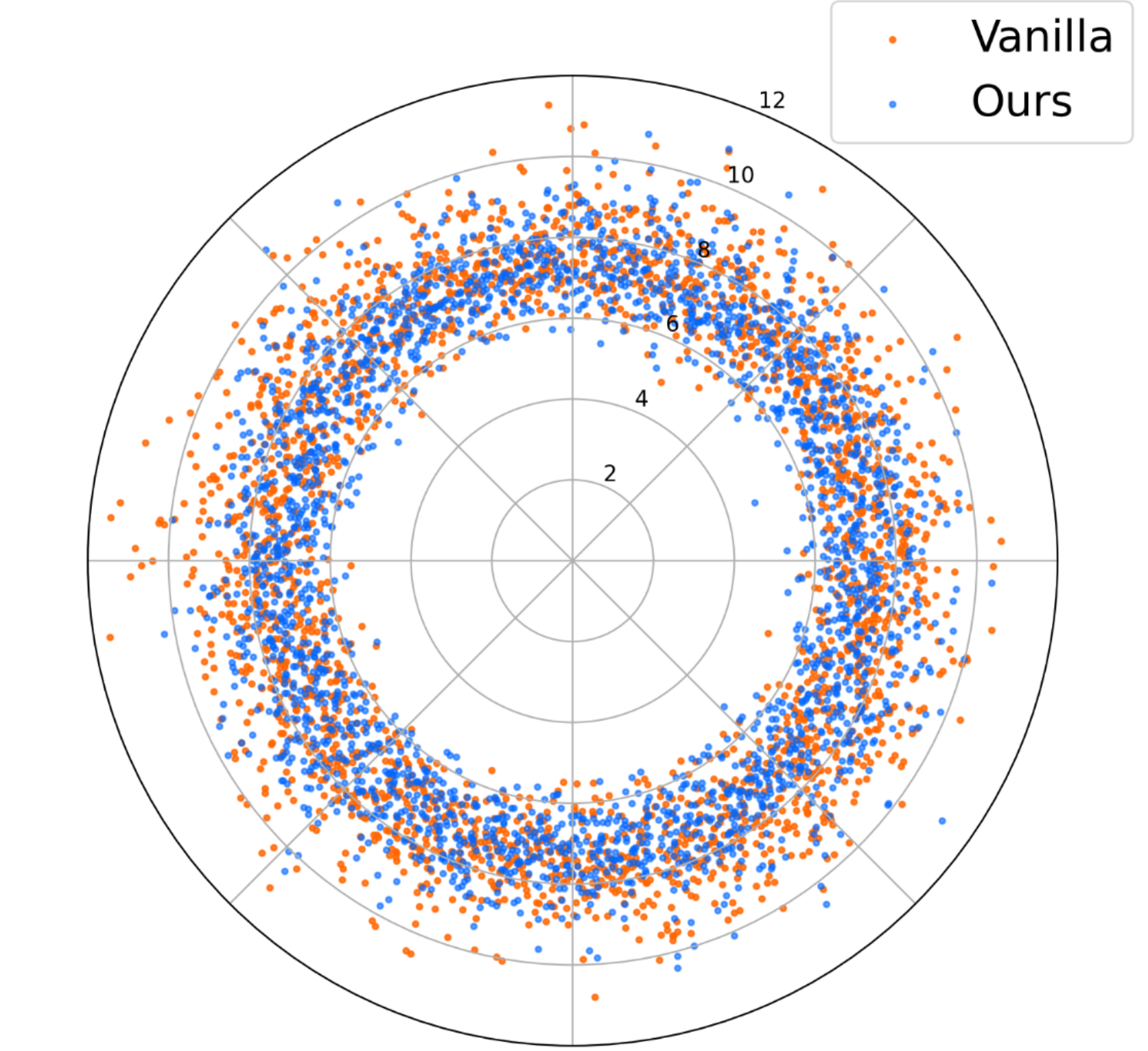}
\figcaption{Statistical variance distribution of patterns learned.}
\label{fig:varition comparison vis weights}
\end{minipage}
\vspace{-6mm}
\end{figure*}

\noindent\textbf{Visualization of Data Distribution.} 
To showcase the efficacy of our algorithm in mitigating model overfitting to particular patterns, we conduct an analysis on the statistical distribution of the patterns learned by the model. 
Specifically, we compute the variance of patterns learned from Level 2 Jitter within the ModelNet-C dataset, as depicted in Fig.~\ref{fig:varition comparison vis weights}. Each point in the figure corresponds to the variance from an individual sample from the dataset, the radius represents the value. Analysis reveals that the pattern distribution associated with the vanilla dropout approach tends to exhibit increased variance, which suggests a propensity for the model to overfit to certain patterns. 
In contrast, APCT prompts the model to integrate information from a broader array of tokens.

\noindent\textbf{Effect of selection number $k$.}
We next investigate the impact of the selection number $k$ of Adversarial Significance Identifier in Fig.~\ref{tab: ablation: k number}. Here $k=1$ means directly treating the max-feat token as the single essential part. This baseline ($k=1$) obtains $75.1\%$ mCE on ModelNet-C. After more essential pattern mining, we observe improvement against corruption, $e.g.$, $75.1\%\rightarrow72.2\%$ mCE when $k=2$. When $k>2$, further increasing $k$ gives marginal performance gains even worse results. We speculate this is because the model is distracted by some trivial patterns due to over-mining.

\noindent\textbf{Effect of mapping rate \textbf{$\gamma$}.}
Tab.~\ref{tab: ablation: mapping ratio} gives the performance with regard to the ratio \textbf{$\gamma$} in mapping function (in Eq.~\ref{eq: mapping function}).
The model performs better with a medium ratio $[0.2, 0.2, 0.2]$ , showing that moderate mapping ratio is more favored. Moreover, at the asymptotic cases of $[\gamma_1, \gamma_2, \gamma_3] = [0.1, 0.2, 0.3]$ or $[\gamma_1, \gamma_2, \gamma_3] = [0.3, 0.2, 0.1]$, the performance drops considerably, evidencing that gradually adjusting the dropping strategy is not a sound solution.

\noindent\textbf{Disparity between ours and vanilla dropout.} 
Our method diverges distinctly from the vanilla dropout, from the perspectives of motivation, technical implementation, and empirical outcomes. 
\ding{182}~From Motivation: we aim to address a nuanced challenge overlooked by standard dropout method: the tendency of deep learning models to make misguided inferences from local features within perturbed samples, particularly prevalent in real-world data scenarios.
\ding{183}~From Implementation: dropout fosters generalization by randomly attenuating network parameters. In contrast, our approach eschews simplistic random dropout in favor of a meticulously engineered mechanism that identifies and diminishes tokens deemed pivotal by the model in its preliminary assessments, thereby compelling the network to recalibrate and shift its focus to previously overlooked tokens, as seen in Fig.~\ref{fig:adversarial_drop}\textcolor{linkcolor}{a} and Fig.~\ref{fig:fullcomparison vis weights}.
\ding{184}~For results, curves in Fig.~\ref{fig:comparison curves of apct and vanilla dropout} demonstrates that when contending with corruptions in the ModelNet-C dataset, our method significantly outperforms the standard dropout ($\textbf{72.2\%}$ vs $74.5\%$).

\vspace{-4mm}
\section{Conclusion}
\label{conclusion}
\vspace{-4mm}
In this work, we introduce a new algorithm, tailored for point cloud recognition in the presence of real-world corruptions. Our algorithm incorporates an adversarial dropping strategy, facilitating the capture of diverse patterns, enabling the assimilate information from non-corrupted regions, thereby ensuring robust prediction under local pattern damaged scenario. Experimental evaluations on comprehensive benchmarks manifest its superiority.

\noindent \textit{Limitations.} While current method can discern different patterns through adversarial strategies and render accurate judgments under corrupted scenarios, the optimal utilization of these cues has not been extensively explored in this paper. In future work, we intend to delve deeper into this aspect.

\clearpage

\noindent\textbf{Acknowledgments}
This work was financially supported by the National Natural Science Foundation of China (No. 62101032), the Postdoctoral Science Foundation of China (Nos. 2021M690015, 2022T150050), and Beijing Institute of Technology Research Fund Program for Young Scholars (No. 3040011182111).

\bibliographystyle{plain}
\bibliography{APCT_submission}


\clearpage
\appendix

\section{Appendix / supplemental material}
The supplementary material herein extends the discussion and analysis presented in the primary manuscript. It is structured as follows:

\textbf{Training Setup Details.}
(\S~\ref{para: Training Setup Details})
: This section delves into the details of our training methodology, offering a comprehensive understanding of our training process.

\textbf{Computational Overhead.} (\S~\ref{para: Computational Overhead}): This section introduces the impact of our method on computational overhead.

\textbf{Results on Clean dataset.} (\S~\ref{para: Results on Clean Dataset}) This section introduces the performance of our method on uncorrupted point cloud dataset.

\textbf{Results on Point Cloud Attack Defense.} (\S~\ref{para: Results on Point Cloud Attack Defense}) This section introduces the ability of our method on point cloud attack defense.

\textbf{Results on MVImageNet.} (\S~\ref{para: mvimagenet results in sub}) This section introduces the performance of our method on more challenging real-world scanned dataset MVImageNet. 

\textbf{Performance of APCT with augmentation methods.} (\S~\ref{para: Performance of APCT with augmentation methods in sub}) This section introduces the performance of our method with various augmentation methods. 

\textbf{Effect of the adversarial mechanism with different tokenization methods.} (\S~\ref{para: Effect of the adversarial mechanism across different tokenization methods in sub}) This section introduce the effect of our adversarial mechanism with different tokenization methods.

\textbf{Effect of the adversarial mechanism incorporating to advanced methods.} (\S~\ref{para: Effect of the adversarial mechanism incorporating to advanced methods in sub}) This section introduce the effect of our adversarial mechanism incorporating to advanced methods.

\textbf{Corruption Implementation Details.}
(\S~\ref{para: corruption details})
: Detailed explication of the corruption settings employed within benchmark datasets is provided, elucidating the experimental conditions and variables.

\textbf{Focal Tokens Identification Implementation.}
(\S~\ref{para: Implementation of focal tokens identification})
: The algorithm for identifying focal tokens is articulated, including a pseudo-code representation in a Pytorch-like syntax, to facilitate replicability and clarity in implementation.

\textbf{Comparative Analysis of Confusion Matrices.} 
(\S~\ref{para:confusion matrix comparison})
: This section presents a comparative study of the confusion matrices resulting from standard transformer-based point cloud models vis-à-vis our proposed algorithm, thereby highlighting the distinctive attributes and performance metrics.

\textbf{More Visualizations of Learned Patterns.} 
(\S~\ref{para:Visualization of learned patterns})
: Further visual evidence is furnished, showcasing the patterns discerned by the classifier's final self-attention layer, thus reinforcing the interpretability aspect of our model.

\textbf{Extended Results.}
(\S~\ref{para: full results in sub})
: Comprehensive results pertaining to ModelNet-C, ScanObjectNN-C, and ShapeNet-C are presented, augmenting the main findings with additional experimental results.

\subsection{Training Setup Details}
\label{para: Training Setup Details}
\noindent\textbf{ModelNet-C.}
Our model is optimized using the AdamW optimizer~\cite{kingma2014adam} with a batch size of 32 for 300 epochs.
The optimizer utilizes a learning rate of 0.0005 and weight decay of 0.05. The CosineAnnealingLR~\cite{loshchilov2016sgdr} scheduler is employed to decrease the learning rate to the minimum value of 1e-6, and the warm up epochs is set to 10. 
On ModelNet, input point cloud is partitioned into 64 tokens with 384 dimension of feature.

\noindent\textbf{ScanObjectNN-C.}
On ScanObjectNN, our model is trained with a batch size of 32 using the AdamW optimizer for 300 epochs. The optimizer is configured with a learning rate of 0.0005 and weight decay of 0.05. The learning rate is decayed to the minimum value of 1e-6 using the CosineLRScheduler scheme, with warp up epochs of 10. 
For the representation, the point cloud data is systematically partitioned into 128 tokens, each having a feature dimension of 384.

\noindent\textbf{ShapeNet-C.}
In this experiment, the proposed model is trained for 300 epochs using the AdamW optimizer. A batch size of 16 is set for the training process. The optimizer was configured with a learning rate of 0.0002 and weight decay of 0.05. Additionally, the learning rate was decayed to the minimum value of 1e-6 by employing the CosineLRScheduler scheme.

\noindent\textbf{Evaluation metrics.} We used evaluation metrics
(mCE, mOA and RmCE) as specified in previous work RPC~\cite{ren2022benchmarking}.

\subsection{Computational Overhead}
\label{para: Computational Overhead}
To demonstrate the computational impact of our approach, we also conducted corresponding comparisons, experiments are conducted on one GeForce RTX 3090. Its impact on computational overhead is presented in Tab.~\ref{tab: ablation: Effect of computational overhead}, including deliberations on both training and inference speed. Noteworthy is that our method stands out for its exemplary memory efficiency, as it refrains from incurring any additional memory overhead. When juxtaposed with the baseline technique, our method, albeit registering a marginal deceleration in the realms of training ($\sim$ 32 samples/s, $6\%$ delay) and inference ($\sim$ 66 samples/s, $6\%$ delay), emerges superior with a pronounced decrement in mCE, registering a downturn of $4.0\%$.

\begin{table}[h]
    \centering
    \setlength{\tabcolsep}{10pt}
    \renewcommand\arraystretch{1.0}
    \caption{Analysis of computational overhead.}
    \label{tab: ablation: Effect of computational overhead}
    \footnotesize
    \begin{tabular}{l c c c c}
    \toprule
    \multirow{2}{*}{Method} & Memory & Train speed & Infer speed & mCE\\
          &  (G)      & (samples/s) & (samples/s) & ($\%$)\\
    \midrule
    Baseline & 10.9  & 415.2 & 1111.7 & 76.2\\
    \color{updatagrey}{~\textbf{$+Ours$}} & 10.9 & 383.4 & 1045.8 & \textbf{72.2}\\
    \midrule
    \color{updatagrey}{~\textbf{$\Delta$}} & \color{updatagrey}{~\textbf{-}} &  \color{updatagrey}{\textsl{$\downarrow\sim$\textbf{6\%}}}  & \color{updatagrey}{\textsl{$\downarrow\sim$\textbf{6\%}}} & \color{updatagrey}{\textsl{$\downarrow$\textbf{4.0}}}  \\
    \bottomrule
    \end{tabular}
\end{table}

\subsection{Results on Clean Dataset}
\label{para: Results on Clean Dataset}
In pursuit to rigorously evaluate and validate the potency of our proposed technique, we also conduct comprehensive evaluations on a pristine dataset.
Specifically, we chose the ScanObjectNN's~\cite{uy2019revisiting} most challenging variant, PB-T50-RS, for our experiments. 
Our proposed technique consistently exhibited improvements even on such clean datasets, a testament underscored by our exhaustive experimental validation. The results presented in Tab.~\ref{tab: ablation: results on clean data} show a discernible performance boost, moving from $85.3\%$ to $86.2\%$. 
Notably, this is a highly competitive result compared with many advanced methods~\cite{qianpointnext, ma2021rethinking}, demonstrating the generalizability of our proposed algorithm. 

\begin{table}[h]
    \centering
    \setlength{\tabcolsep}{20pt}
    \caption{Results on ScanObjectNN, OA($\%, \uparrow$) is reported. Note: No multi-scale inference is employed. Baseline denotes model w/o Adversarial Dropping. }
    \label{tab: ablation: results on clean data}
    \footnotesize
    \begin{tabular}{l c}
    \toprule
    Method  & OA  \\
    \midrule
    DGCNN~\cite{wang2019dynamic}  & 78.1 \\ 
    PointNet~\cite{qi2017pointnet} & 68.2 \\ 
    PointNet$++$~\cite{qi2017pointnet++}  & 77.9 \\ 
    PointMLP~\cite{ma2021rethinking}  & 85.4 \\
    RPC~\cite{ren2022benchmarking}  & 74.7 \\
    PointNeXt~\cite{qianpointnext} & \textbf{87.3} \\
    \midrule
    Baseline & 85.3\\
    + Ours & 86.2~\textcolor{PineGreen!60}{$\uparrow$0.9} \\
    \bottomrule
    \end{tabular}
\end{table}

\subsection{Results on Point Cloud Attack Defense}
\label{para: Results on Point Cloud Attack Defense}
\noindent\textbf{Setup.}
Point cloud attack for classification aims to manipulate the input point cloud in a manner that induces misclassification by a well-tuned classifier. Encouraged by the remarkable performance of our algorithm against corruption, we investigate its potential application in point cloud defense. To evaluate this, we assess the performance of baseline model trained using our method on various attacks, including point perturbation attack~\cite{xiang2019generating}, individual point adding attack~\cite{xiang2019generating}, kNN attack~\cite{tsai2020robust}, and point dropping attack~\cite{zheng2019pointcloud}.
Consistent with previous works~\cite{wu2020if,zhou2019dup}, we conduct targeted attacks and report the resulting classification accuracy.
Higher accuracy indicates better defense against attack.

\noindent\textbf{Main Results.}
Based on the results presented in Tab.~\ref{tab: attack results}, it is observed that our algorithm consistently perform well on defending all attacks, especially on two point dropping attacks. The accuracy improvement can be as high as $7.29\%$ in the Drop-200 attack, indicating the scalability and effectiveness of the approach. For point cloud attack defend, our method perform well on defending all attacks, as seen in Tab.~\ref{tab: attack results}. Even with the Perturb attack, which introduces very fine perturbations leading to a chaotic point cloud, APCT achieves a good performance of $74.31\%$ OA. These results evident the strong generalization ability of our APCT. The above analysis leads to the conclusion that our method has a strong generalization ability across various point cloud attack algorithms.

\begin{table}[!t]
    \centering
    \setlength{\tabcolsep}{7.5pt}
    \footnotesize
    \caption{
    Results on point cloud attack defense, OA($\%, \uparrow$).}
    \label{tab: attack results}
    \begin{tabular}{c|cccccc}
        \toprule 
        \textbf{Method} & Perturb & Add-CD & Add-HD & kNN & Drop-100 & Drop-200 \\ 
        \hline
        No Defense & 0.00 & 0.00 & 0.12 & 0.00 & 80.55 & 69.53  \\
        \rowcolor{RowColor}APCT(Ours) & 74.31  & 72.85 & 54.82 & 49.96 & 84.08 & 76.82 \\ 
        \hline
        \fontsize{7.5pt}{1em}\selectfont{\color{updatagrey}{~\textbf{$\Delta$}}} & \updata{+74.31}  & \updata{+72.85} & \updata{+54.70} & \updata{+49.96} & \updata{+3.53} & \updata{+7.29} \\ 
        \bottomrule
    \end{tabular}
\end{table}

\subsection{Results on MVImageNet}
\label{para: mvimagenet results in sub}
We further evaluate the performance of our approach on one additional dataset MVImageNet~\cite{yu2023mvimgnet}. It is a challenging benchmark for real-world point cloud classification, which contains 64,000 training and 16,000 testing samples. As shown in the Tab.~\ref{tab: results on mvimagenet}, our approach achieves $86.6\%$ OA and still exhibits good generalization capacity in real-world scenarios.

\begin{table}[h]
    \centering
    \setlength{\tabcolsep}{20pt}
    \caption{Classification results on the MVIamgeNet dataset, OA($\%, \uparrow$) is reported. }
    \label{tab: results on mvimagenet}
    \footnotesize
    \begin{tabular}{l c}
    \toprule
    Method  & OA ($\%, \uparrow$)  \\
    \midrule
    PointNet~\cite{qi2017pointnet} & 70.7 \\ 
    PointNet++~\cite{qi2017pointnet++}  & 79.2 \\ 
    DGCNN~\cite{wang2019dynamic}  & 86.5 \\ 
    PAConv~\cite{xu2021paconv} & 83.4 \\
    PointMLP~\cite{ma2021rethinking}  & 88.9 \\
    \midrule
    APCT (Ours) & 86.6\\
    \bottomrule
    \end{tabular}
\end{table}

\subsection{Performance of APCT with augmentation methods}
\label{para: Performance of APCT with augmentation methods in sub}
We further evaluate the performance of our approach with various data augmentation methods on ModelNet-C dataset, the results are as follows. Beside the discussed PointMixup~\cite{chen2020pointmixup} and PointCutMix~\cite{zhang2022pointcutmix}, we additionally incorporate experiments with the data augmentation techniques PointWOLF~\cite{kim2021point}, RSMix~\cite{lee2021regularization}, and WOLFMix~\cite{ren2022benchmarking}. 

Among these, PointMixup, PointCutMix and RSMix fall under the category of mixing augmentation, where they mix several point clouds following pre-defined regulations. PointWOLF pertains to deformation techniques that non-rigidly deforms local parts of an object. WOLFMix combines both mixing and deformation augmentations, which first deforms the object, and subsequently rigidly mixes the deformed objects together.

As shown in the Tab.~\ref{tab: results with augmentation}, data augmentation methods further improve the robustness of our method against point cloud corruptions. Employing mixing or deformation data augmentation techniques independently can enhance the robustness of the model, \textit{e.g.}, the results of our model with PointWOLF ($67.0\%$ mCE) and with PointMixup ($66.2\%$ mCE). When these two techniques are combined, as in WOLFMix, the robustness of the model is further augmented ($64.7\%$ mCE). Additionally, these experiments demonstrate the compatibility of our method with various data augmentation techniques, further underscoring its potential in addressing data corruption. 

\begin{table}[h]
    \centering
    \setlength{\tabcolsep}{20pt}
    \caption{Performance of APCT with augmentation methods on ScanObjectNN-C. }
    \label{tab: results with augmentation}
    \footnotesize
    \begin{tabular}{l c}
    \toprule
    Method  & mCE ($\%, \downarrow$)  \\
    \midrule
    APCT (Ours) & 72.2\\
    + PointMixup~\cite{chen2020pointmixup} & 66.2 \\
    + PointCutMix-R~\cite{zhang2022pointcutmix} & 69.7 \\
    + PointWOLF~\cite{kim2021point} & 67.0 \\
    + RSMix~\cite{lee2021regularization} & 71.3 \\
    + WOLFMix~\cite{ren2022benchmarking} & 64.7 \\
    \bottomrule
    \end{tabular}
\end{table}

\subsection{Effect of the adversarial mechanism with different tokenization methods}
\label{para: Effect of the adversarial mechanism across different tokenization methods in sub}

To evaluate the generalization and robustness of our method, we apply our method with two alternative tokenization methods: mini-DGCNN and mini-PCT. The results are shown in the Tab.~\ref{tab: results of the adversarial mechanism across different tokenization methods}.

As we can see from the results, the performance of our adversarial mechanism is relatively stable across different tokenization methods. This indicates that our method is not sensitive to the specific tokenization method used and can effectively improve the robustness of point cloud models.

\begin{table}[t]
    \centering
    \setlength{\tabcolsep}{20pt}
    \caption{Results of the adversarial mechanism with different tokenization methods. }
    \label{tab: results of the adversarial mechanism across different tokenization methods}
    \footnotesize
    \begin{tabular}{l c}
    \toprule
    Method  & mCE ($\%, \downarrow$)  \\
    \midrule
    mini-DGCNN & 71.1 \\ 
    mini-PCT & 72.4 \\ 
    mini-PointNet (Ours)  & 72.2 \\ 
    \bottomrule
    \end{tabular}
\end{table}

\subsection{Effect of the adversarial mechanism incorporating to advanced methods}
\label{para: Effect of the adversarial mechanism incorporating to advanced methods in sub}

We have extended the adversarial 'digging sub-optimal patterns' mechanism to two state-of-the-art (SOTA) methods, PointM2AE~\cite{zhang2022pointm2ae} and PointGPT~\cite{chen2024pointgpt}, on the ModelNet-C dataset. The results are promising and demonstrate the general applicability of our approach.

As shown in the Tab.~\ref{tab: results of the adversarial mechanism incorporating to advanced methods} below, incorporating our 'digging sub-optimal patterns' mechanism into PointM2AE and PointGPT resulted in significant reduction in mCE scores. These results suggest that our approach can effectively enhance the robustness of various point cloud recognition models. By encouraging the model to explore and utilize a broader range of patterns, our method enables the models to better generalize to corrupted data.

\begin{table}[h]
    \centering
    \setlength{\tabcolsep}{20pt}
    \renewcommand\arraystretch{0.75}
    \caption{Results of the adversarial mechanism incorporating to advanced methods. }
    \label{tab: results of the adversarial mechanism incorporating to advanced methods}
    \footnotesize
    \begin{tabular}{l c}
    \toprule
    Method  & mCE ($\%, \downarrow$)  \\
    \midrule
    PointM2AE & ~~ 83.9 \\ 
    
    \color{updatagrey}{~\textbf{$+Ours$}} & 82.9  \color{updatagrey}{\textsl{$\downarrow$\textbf{$1.0\%$}}} \\
    PointGPT & ~~ 83.4 \\ 
    \color{updatagrey}{~\textbf{$+Ours$}} & 82.0  \color{updatagrey}{\textsl{$\downarrow$\textbf{$1.4\%$}}} \\
    \bottomrule
    \end{tabular}
\end{table}

\begin{table}[t]
\centering
\setlength{\abovecaptionskip}{2pt}  
\setlength{\belowcaptionskip}{2pt}
\renewcommand{\arraystretch}{0.9}
\setlength{\tabcolsep}{12.0pt}
    \caption{Point cloud corruption benchmarks comparison.}
    \label{tab:Point cloud corruption benchmarks comparison}
    \begin{tabular}{lccc}
        \toprule 
        Dataset & Refernce & Number Samples & Real-world \\
        \midrule
        ModelNet-C~\cite{ren2022benchmarking} & ICML2022 & 2468 & \ding{55}\\
        ShapeNet-C~\cite{ren2022pointcloud} & Arxiv2022 & 2874 & \ding{55}\\
        ScanObjectNN-C~\cite{wang2023adaptpoint} & ICCV2023 & 2882 & \ding{51}\\
        \bottomrule
    \end{tabular}
\end{table}


\subsection{Corruption Implementation Details}
\label{para: corruption details}
\begin{figure}[t]
\centering
\setlength{\abovecaptionskip}{2pt}  
\setlength{\belowcaptionskip}{2pt}
\includegraphics[width=1.0\columnwidth]{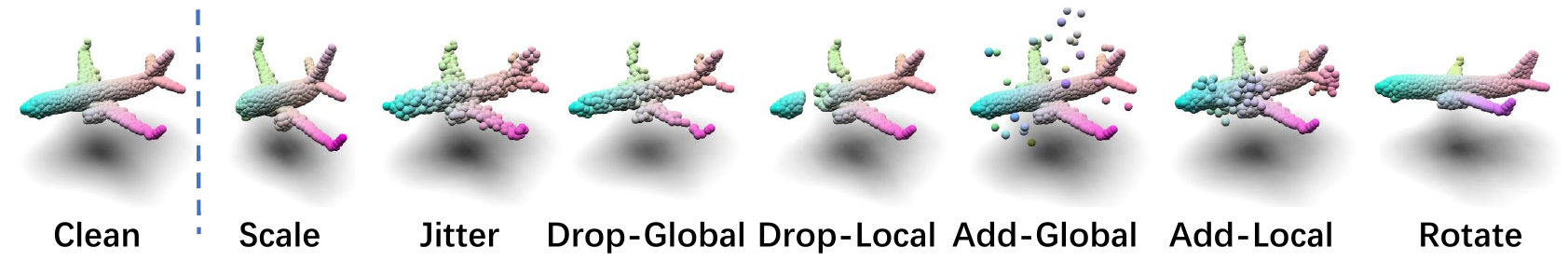} 
\caption{Visualization of samples in ModelNet-C, which is constructed by seven types of corruptions with five levels of severity. Listed examples are from severity level 2.}
\label{fig:modelnetc_sample}
\end{figure}

In the realm of corruption benchmarks, ModelNet-C~\cite{ren2022benchmarking}, ScanObjectNN-C~\cite{wang2023adaptpoint}, and ShapeNet-C~\cite{ren2022pointcloud} are prominent datasets, as detailed in Tab.~\ref{tab:Point cloud corruption benchmarks comparison}. Among these, ScanObjectNN-C is particularly noteworthy for its real-world scanned dataset, presenting a heightened level of challenge. 
Central to the analysis of these benchmarks is the application of seven atomic corruptions, which serve as fundamental perturbations for evaluating algorithmic resilience in the face of data degradation. Details about the corruptions are as follows:
\begin{itemize}
  \item \textit{Scale}: Application of random anisotropic scaling to the point cloud.
  \item \textit{Rotate}: Rotation of the point cloud by a small angle.
  \item \textit{Jitter}: Addition of Gaussian noise to point coordinates.
  \item \textit{Drop-Global}: Random removal of points from the point cloud.
  \item \textit{Drop-Local}: Random elimination of several local clusters from the point cloud.
  \item \textit{Add-Global}: Addition of random points sampled within a unit sphere. 
  \item \textit{Add-Local}: Expansion of random points on the point cloud into normally distributed clusters.
\end{itemize}
Each corruption type is associated with five severity levels, facilitating a comprehensive evaluation of robustness. This diverse set of corruptions serves to systematically assess the resilience of models against various perturbations in the point cloud data.
The differential attributes between samples subjected to an array of corruptions and the clean sample are visually delineated in Fig.~\ref{fig:modelnetc_sample}. An observable shift in structural fidelity is evident in the corrupted instances when juxtaposed against their unaltered originals. These perturbations markedly hinder the performance efficacy of computational models, culminating in a consistent decline in model accuracy across a spectrum of corruption categories.

\subsection{Focal Tokens Identification Implementation}
\label{para: Implementation of focal tokens identification}
Algorithm.~\ref{pserudo-code: identify focal tokens} shows the implementation of focal tokens identification introduced in the main manuscript. Through such implementation, we can easily calculate all token significances associated to overall perception and identify the focal ones.

\begin{algorithm}[!h]
\centering
\setlength{\tabcolsep}{2.0pt}
\renewcommand\arraystretch{0.5}
\setlength{\abovecaptionskip}{-2pt}  
\setlength{\belowcaptionskip}{2pt}
\caption{Pseudo-Code of \textit{Identifying focal tokens} in a Pytorch-like Style.} 
\begin{lstlisting}[language={Python}]
`{\textbf{Input}}:`
  tokens - [N, C], where N denotes the length `and` C denotes the feat dimensions.
  k - number of focal tokens selected `in` each feat channel, default 2.
`{\textbf{Output}}:`
  matrix - [N], each value within matrix enumerates number within the `range` (0,C), indicating the token significance associated to overall perception.

# identify focal tokens
`{\textbf{Function}}:` identify_focaltokens(tokens, k):
    N, C = tokens.shape
    # sort token in each feat channel
    _, sort_idx = sort(tokens, dim=0, descending=True)
    # select the idxs of top k tokens 
    idx_topk = sort_idx[:k, :]           
    idx_topk = idx_topk.view(-1)   
    matrix = zeros(N)
    # accumulate the idxs
    matrix.scatter_add(1, idx_topk, ones_like(idx_topk))
    `return` matrix
\end{lstlisting}
\label{pserudo-code: identify focal tokens}
\end{algorithm}

\subsection{Comparative Analysis of Confusion Matrices}
\label{para:confusion matrix comparison}
\begin{figure*}[h]
\centering
\setlength{\abovecaptionskip}{2pt}  
\setlength{\belowcaptionskip}{2pt}
  \includegraphics[width=0.98\textwidth]{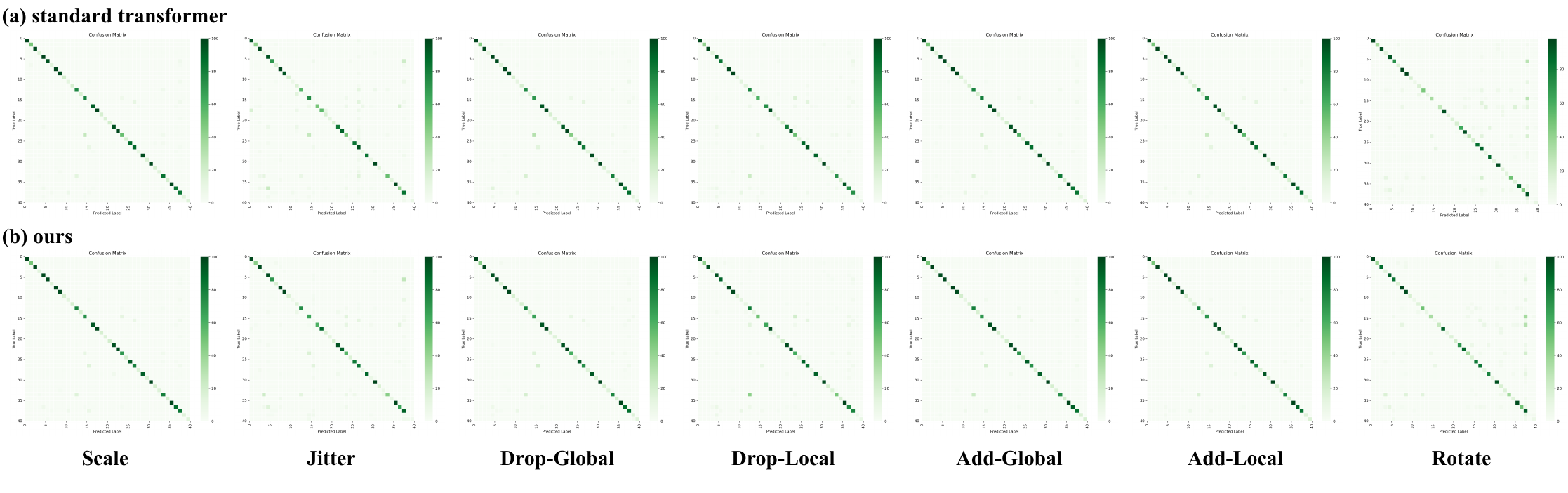}
  \caption{Side-by-side comparison of the confusion matrices for two distinct architectures: the standard transformer-based point cloud model (top row) and ours (bottom row).
  }
  \label{fig:fullcomparison confusion matrix}
\end{figure*}

To ascertain the efficacy of our method on corrupted data, we selected all corruption types at an intermediate level (level 2) from ModelNet-C~\cite{ren2022benchmarking} for comparative experiments. As illustrated in Fig.~\ref{fig:fullcomparison confusion matrix}, we present side-by-side confusion matrix comparisons. Notably, our adversarial dropping strategy substantially enhances the model's performance under corruption. This suggests that our approach incentivizes the model to delve into a broader spectrum of patterns, ultimately converging to a global motif. Consequently, even if specific local motifs deteriorate within corrupted data, the model retains the capability to glean information from alternative regions for proficient predictions.

\subsection{More Visualizations of learned patterns}
\label{para:Visualization of learned patterns}
\begin{figure*}[h]
\centering
\setlength{\abovecaptionskip}{2pt}  
\setlength{\belowcaptionskip}{2pt}
  \includegraphics[width=0.98\textwidth]{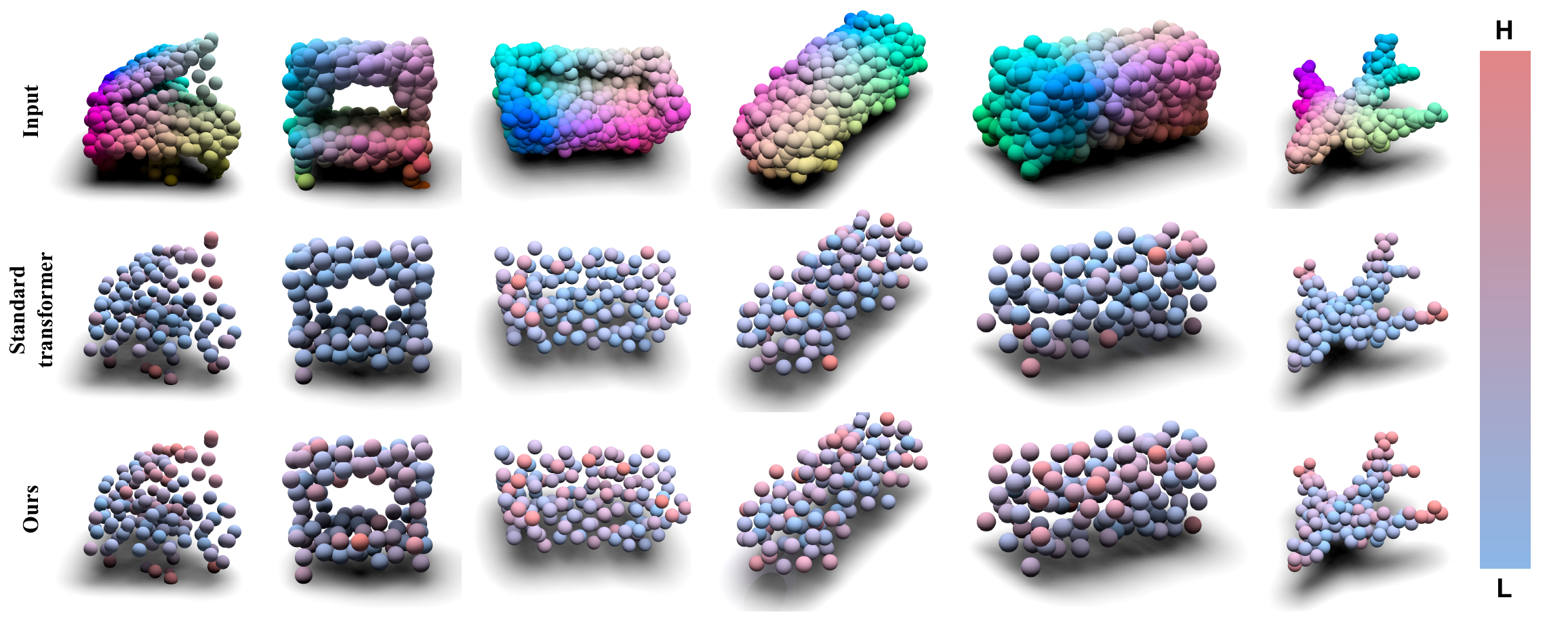}
  \caption{More visualization results of patterns learned by the classifier on ModelNet-C. Tokens with \textcolor{linkcolor}{high} / \textcolor{citecolor}{low} scores are in \textcolor{linkcolor}{red} / \textcolor{citecolor}{blue}, respectively. }
  \label{fig:fullcomparison vis weights full}
\vspace{-5mm}
\end{figure*}

To elucidate the implications of our adversarial dropping strategy on real-world corruptions, we embarked on a methodical analysis, juxtaposing token features that the classifier attends to, particularly on the perturbed data from the jitter-2 test suite of ModelNet-C. Fig.~\ref{fig:fullcomparison vis weights full} offers a visual delineation of the classifier's final self-attention layer. Specifically, we adopt the \textit{Focal tokens identification} procedure, as outlined in the main manuscript, post-normalization. The computed token features are then harnessed to epitomize their significance to the model. Tokens rendering substantial contributions to the perception model are depicted with heightened color intensities: red symbolizing high contribution, while blue denotes low contribution. 
The visualizations indicate that, standard architectures tend to overfit to specific local patterns, such as table legs or airplane wings. In contrast, our approach fosters a more comprehensive exploration of patterns, culminating in the capture of a global motif. Hence, even if some local motifs degrade in corrupted data, the model can still extract valuable information from other regions for effective prediction.

\subsection{Extended Results}
\label{para: full results in sub}
\noindent\textbf{ModelNet-C.}
We present complete results for both the mOA and RmCE metrics in Tab.~\ref{tab:modelnet40-c, mOA} and Tab.~\ref{tab:modelnet40-c, RmCE}, respectively.

\noindent\textbf{ScanObjectNN-C.}
We show full results for the mOA metric and the RmCE metrics in Tab.~\ref{tab:ScanObjectNN-c, mOA} and Tab.~\ref{tab:ScanObjectNN-c, RmCE}, respectively.

\noindent\textbf{ShapeNet-C.}
The complete results of the RmCE metric are presented in Tab.~\ref{tab:results_shapenet-c, RmCE} for ShapeNet-C dataset.

\begin{table*}[h]
    \centering
    \setlength{\tabcolsep}{6.0pt}
    \renewcommand\arraystretch{0.9}
    \setlength{\abovecaptionskip}{-2pt}  
    \setlength{\belowcaptionskip}{2pt}
    \caption{Classification results on the ModelNet-C dataset, mOA($\%, \uparrow$) is reported, the best performance is \textbf{bold}.  $\dagger$ denotes method designed specifically against corruption.}
    \label{tab:modelnet40-c, mOA}
    \footnotesize
    \begin{tabular}{l|c|cccccccc}
        \toprule 
        Method  & ~mOA~  & Scale & Jitter & Drop-G & Drop-L & Add-G & Add-L & Rotate \\ 
        \midrule
        DGCNN\pub{TOG2019}~\cite{wang2019dynamic}  & 76.4  & 90.6  & 68.4  & 75.2  & 79.3  & 70.5  & 72.5  & 78.5  \\ %
        PointNet\pub{CVPR2017}~\cite{qi2017pointnet} & 65.8  & 88.1  & {79.7}  & 87.6  & 77.8  & 12.1  & 56.2  & 59.1  \\  %
        PointNet++\pub{NIPS2017}~\cite{qi2017pointnet++}  &  75.1  & 91.8  & 62.8  & 84.1  & 62.7  & 81.9  & 72.7  & 69.8   \\ %
        GDANet\pub{AAAI2021}~\cite{xu2021learning}  & 78.9  & {92.2}  & 73.5  & 80.3  & 81.5  & 74.3  & 71.5  & 78.9    \\ %
        PAConv\pub{CVPR2021}~\cite{xu2021paconv} & 73.0  & 91.5  & 53.7  & 75.2  & 79.2  & 68.0  & 64.3  & {79.2}  \\ %
        PCT\pub{CVM2021}~\cite{guo2021pct} & 78.1  & 91.8  & 72.5  & 86.9  & 79.3  & 77.0  & 61.9  & 77.6   \\ %
        RPC$^{\dagger}$\pub{ICML2022}~\cite{ren2022benchmarking} & 79.5  & 92.1  & 71.8  & 87.8  & {83.5}  & 72.6  & 72.2  & 76.8   \\ %
        PointNeXt\pub{NIPS2022}~\cite{qianpointnext} & 80.5  & 91.5  & 59.0  & 79.0  & 80.2  & {92.6}  & {92.4}  & 68.6 \\%
        PointM2AE\pub{NIPS2022}~\cite{zhang2022pointm2ae} & 80.6 & 91.0	& 49.7 & 	86.8 &	82.7 &	91.1 &	91.6 &	71.0 \\
        PointGPT\pub{NIPS2023}~\cite{chen2024pointgpt} & 81.7 &	89.8 &	67.6 &	85.5 &	79.7 &	91.4 &	91.0 &	66.8\\
        
        \midrule
        \rowcolor{RowColor} APCT (Ours) & \textbf{84.1}  & 91.1  & 72.1  & {88.4}  & 82.4  & 91.6  & 91.8  & 71.5  \\
        \fontsize{7.5pt}{1em}\selectfont{\color{updatagrey}{\textit{~{vs. prev. SoTA}}}} & \fontsize{7.5pt}{1em}\selectfont{\color{updatagrey}{\textsl{$\uparrow$\textbf{2.4}}}}  & \multicolumn{7}{c}{\ \ \ - \ \ \ } \\
        \bottomrule
    \end{tabular}
\end{table*}

\begin{table*}[h]
    \centering
    \setlength{\tabcolsep}{6.0pt}
    \renewcommand\arraystretch{0.9}
    \setlength{\abovecaptionskip}{-2pt}  
    \setlength{\belowcaptionskip}{2pt}
    \caption{Classification results on the ModelNet-C dataset, RmCE($\%, \downarrow$) is reported, the best performance is \textbf{bold}.  $\dagger$ denotes method  specifical against corruption.}
    \label{tab:modelnet40-c, RmCE}
    \footnotesize
    \begin{tabular}{l|c|cccccccc}
        \toprule 
        Method  & ~RmCE~  & Scale & Jitter & Drop-G & Drop-L & Add-G & Add-L & Rotate \\ 
        \midrule
        DGCNN\pub{TOG2019}~\cite{wang2019dynamic}  & 100.0  & 100.0  & 100.0  & 100.0  & 100.0  & 100.0  & 100.0  & {100.0}  \\
        PointNet\pub{CVPR2017}~\cite{qi2017pointnet} & 148.8  & 130.0  & {45.5}  & {17.8}  & 97.0  & 355.7  & 171.6  & 224.1  \\ 
        PointNet++\pub{NIPS2017}~\cite{qi2017pointnet++}  & 111.4  & 60.0  & 124.8  & 51.1  & 227.8  & 50.2  & 101.0  & 164.5  \\ 
        GDANet\pub{AAAI2021}~\cite{xu2021learning}  & 86.5  & 60.0  & 82.2  & 75.3  & 89.5  & 86.4  & 109.0  & 102.8     \\ 
        PAConv\pub{CVPR2021}~\cite{xu2021paconv} & 121.1  & 105.0  & 164.9  & 105.7  & 108.3  & 115.8  & 145.8  & 102.1  \\ 
        PCT\pub{CVM2021}~\cite{guo2021pct} & 88.4  & 60.0  & 84.7  & 35.1  & 103.0  & 72.4  & 154.7  & 109.2    \\ 
        RPC$^{\dagger}$\pub{ICML2022}~\cite{ren2022benchmarking} & 77.8  & 45.0  & 87.6  & 29.9  & {71.4}  & 92.3  & 103.5  & 114.9     \\ 
        PointNeXt\pub{NIPS2022}~\cite{qianpointnext} & 76.6  & 55.0  & 138.8  & 78.2  & 93.2  & {0.0}  & 1.0  & 170.2 \\
        PointM2AE\pub{NIPS2022}~\cite{zhang2022pointm2ae} & 57.7 &	5.0 &	171.1 &	24.7 &	63.2 &	0.0 &	-2.5 &	142.6 \\
        PointGPT\pub{NIPS2023}~\cite{chen2024pointgpt} & 68.1 &	80.0 &	98.3  &	33.9 &	88.0 &	0.0 &	2.0 &	174.5\\
        \midrule
        \rowcolor{RowColor} APCT (Ours) & \textbf{51.4}  & {40.0}  & 81.8  & 20.1  & {71.4}  & 1.4  & {0.5}  & 144.7 \\
        \fontsize{7.5pt}{1em}\selectfont{\color{updatagrey}{\textit{~{vs. prev. SoTA}}}} & \fontsize{7.5pt}{1em}\selectfont{\color{updatagrey}{\textsl{$\downarrow$\textbf{6.3}}}}  & \multicolumn{7}{c}{\ \ \ - \ \ \ } \\
        \bottomrule
    \end{tabular}
\end{table*}

\clearpage

\begin{table}[!t]
    \centering
    \setlength{\tabcolsep}{6.0pt}
    \renewcommand\arraystretch{0.9}
    \setlength{\abovecaptionskip}{-2pt}  
    \setlength{\belowcaptionskip}{2pt}
    \caption{Classification results on the ScanObjectNN-C dataset, mOA($\%, \uparrow$) is reported.  $\dagger$ denotes method designed specifically against corruption.}
    \label{tab:ScanObjectNN-c, mOA}
    \footnotesize
    \begin{tabular}{l|c|cccccccc}
        \toprule 
        Method & ~mOA~ & Scale & Jitter & Drop-G & Drop-L & Add-G & Add-L & Rotate \\ 
        \midrule
        DGCNN\pub{TOG2019}~\cite{wang2019dynamic} & 62.8  & 57.8  & 45.6  & 62.2  & 69.7  & 54.0  & 77.3  & 73.3  \\ 
        PointNet\pub{CVPR2017}~\cite{qi2017pointnet} & 53.3  & 40.4  & {51.9}  & 70.3  & 62.0  & 35.9  & 54.6  & 58.1 \\ 
        PointNet++\pub{NIPS2017}~\cite{qi2017pointnet++} &  64.1  & 62.1  & 40.0  & 79.2  & 61.3  & 56.4  & 79.5  & 70.5  \\ 
        RPC$^{\dagger}$\pub{ICML2022}~\cite{ren2022benchmarking} &  52.2  & 44.4  & 41.6  & 44.9  & 60.4  & 47.4  & 64.0  & 62.6   \\
        PointNeXt\pub{NIPS2022}~\cite{qianpointnext} & 65.5  & {66.1}  & 41.3  & 69.5  & 71.4  & 56.5  & 80.1  & {73.4}   \\
        PointM2AE\pub{NIPS2022}~\cite{zhang2022pointm2ae} & 71.0 &	71.1 &	35.1 & 79.1	& 74.8 &	83.6 &	83.8 & 69.5\\
        PointGPT\pub{NIPS2023}~\cite{chen2024pointgpt} & 68.7 &	49.4 &	43.4 & 77.9 &	76.7 &	83.4 &	83.6 & 66.6 \\
        
        \midrule
        \rowcolor{RowColor} APCT (Ours) &\textbf{72.4}  & 55.9  & 46.2  & {81.5}  & {79.7}  & {86.2}  & {85.7}  & 71.4  \\
        \fontsize{7.5pt}{1em}\selectfont{\color{updatagrey}{\textit{~{vs. prev. SoTA}}}} & \fontsize{7.5pt}{1em}\selectfont{\color{updatagrey}{\textsl{$\uparrow$\textbf{1.4}}}}  & \multicolumn{7}{c}{\ \ \ - \ \ \ } \\
        \bottomrule
    \end{tabular}
\end{table}

\begin{table}[!t]
    \centering
    \setlength{\tabcolsep}{6.0pt}
    \renewcommand\arraystretch{0.9}
    \setlength{\abovecaptionskip}{-2pt}  
    \setlength{\belowcaptionskip}{2pt}
    \caption{Classification results on the ScanObjectNN-C dataset, RmCE($\%, \downarrow$) is reported.  $\dagger$ denotes method designed specifically against corruption.}
    \label{tab:ScanObjectNN-c, RmCE}
    \footnotesize
    \begin{tabular}{l|c|cccccccc}
        \toprule 
        Method & ~RmCE~ & Scale & Jitter & Drop-G & Drop-L & Add-G & Add-L & Rotate \\ 
        \midrule
        DGCNN\pub{TOG2019}~\cite{wang2019dynamic} & 100.0  & 100.0  & 100.0  & 100.0  & 100.0  & 100.0  & 100.0  & 100.0 \\ 
        PointNet\pub{CVPR2017}~\cite{qi2017pointnet} & 106.1  & 120.2  & {55.1}  & {15.8}  & 74.8  & 119.9  & 228.9  & 127.7 \\ 
        PointNet++\pub{NIPS2017}~\cite{qi2017pointnet++} & 97.6  & 85.9  & 114.8  & 29.7  & 154.4  & 93.5  & 79.1  & 125.9 \\ 
        RPC$^{\dagger}$\pub{ICML2022}~\cite{ren2022benchmarking} &  102.1  & 108.2  & 82.3  & 126.0  & 88.8  & 85.8  & 126.6  & {97.3}  \\
        PointNeXt\pub{NIPS2022}~\cite{qianpointnext} & 93.9  & {75.8}  & 114.4  & 75.5  & 98.6  & 96.7  & 84.7  & 111.2   \\
        PointM2AE\pub{NIPS2022}~\cite{zhang2022pointm2ae} & \textbf{49.9} &	44.6 &	120.6 & 19.1 &	54.7 &	0.0 &	0.0 &  112.8\\
        PointGPT\pub{NIPS2023}~\cite{chen2024pointgpt} & 59.7 &	121.4 &	99.5 & 23.3 &	41.6 &	0.0 &	0.0 &  134.4\\
        
        \midrule
        \rowcolor{RowColor} APCT (Ours) & {56.0}  & 108.2  & 99.5  & 19.9  & {40.4}  & {0.0}  & {5.9}  & 118.4 \\
        \fontsize{7.5pt}{1em}\selectfont{\color{updatagrey}{\textit{~{vs. prev. SoTA}}}} & \fontsize{7.5pt}{1em}\selectfont{\color{updatagrey}{\textsl{$\uparrow$\textbf{6.1}}}}  & \multicolumn{7}{c}{\ \ \ - \ \ \ } \\
        \bottomrule
    \end{tabular}
\end{table}

\begin{table}[!t]
    \centering
    \setlength{\tabcolsep}{4.3pt}
    \renewcommand\arraystretch{0.9}
    \setlength{\abovecaptionskip}{-2pt}  
    \setlength{\belowcaptionskip}{2pt}
    \caption{Segmentation results in terms of RmCE ($\%, \downarrow$) on ShapeNet-C dataset.} 
    \label{tab:results_shapenet-c, RmCE}
    \begin{tabular}{l|c|ccccccc}
        \toprule 
        Method & RmCE & Scale & Jitter & Drop-G & Drop-L & Add-G & Add-L & Rotate \\ 
        \midrule
        DGCNN\pub{TOG2019}~\cite{wang2019dynamic} & 100.0  & 100.0  & 100.0  & 100.0  & 100.0  & 100.0  & 100.0  & 100.0  \\
        PointNet\pub{CVPR2017}~\cite{qi2017pointnet} & 105.6  & 35.5  & \textbf{88.0}  & \textbf{8.7}  & 115.2  & 156.6  & 120.6  & 214.4  \\ 
        PointNet++\pub{NIPS2017}~\cite{qi2017pointnet++} &  185.0  & 68.5  & 132.9  & 17.6  & 786.0  & 83.0  & 116.9  & 90.1  \\ 
        PAConv\pub{CVPR2021}~\cite{xu2021paconv} & 84.8  & 56.0  & 133.6  & 76.4  & 78.9  & 59.7  & 94.7  & 94.0  \\ 
        GDANet\pub{AAAI2021}~\cite{xu2021learning} & 78.5  & 26.4  & 111.5  & 80.6  & 84.2  & 53.5  & 95.2  & 97.9  \\
        PT\pub{ICCV2021}~\cite{zhao2021point} & 93.3  & 98.1  & 105.1  & 72.8  & 107.7  & 113.3  & 105.4  & \textbf{50.7}  \\
        PointMLP\pub{ICLR2022}~\cite{ma2021rethinking} & 81.0  & 47.4  & 142.8  & 21.7  & 96.1  & 88.2  & 110.9  & 60.1  \\ 
        Point-BERT\pub{CVPR2022}~\cite{yu2022point_pointbert} & 89.5  & 28.3  & 135.6  & 21.3  & 61.9  & 130.3  & 136.0  & 112.8  \\ 
        Point-MAE\pub{ECCV2022}~\cite{pang2022masked} & 70.3  & \textbf{18.0}  & 120.9  & 22.2  & \textbf{45.9}  & 65.0  & 108.8  & 111.4  \\ 
        \midrule
        \rowcolor{RowColor} APCT (Ours) & \textbf{55.3}	& 39.9	& 137.3 & 20.6 & 70.3 &	\textbf{0.1} & \textbf{0.4} & 118.5 \\
        \fontsize{7.5pt}{1em}\selectfont{\color{updatagrey}{\textit{~{vs. prev. SoTA}}}} & \fontsize{7.5pt}{1em}\selectfont{\color{updatagrey}{\textsl{$\downarrow$\textbf{15.0}}}}  & \multicolumn{7}{c}{\ \ \ - \ \ \ } \\
        \bottomrule
    \end{tabular}
\end{table}

\end{document}